\documentclass[sigconf,dvipsnames,natbib=true,anonymous=False]{acmart}

\AtBeginDocument{%
  \providecommand\BibTeX{{%
    \normalfont B\kern-0.5em{\scshape i\kern-0.25em b}\kern-0.8em\TeX}}}

\copyrightyear{2022}
\acmYear{2022}
\setcopyright{acmcopyright}\acmConference[KDD '22]{Proceedings of the 28th ACM
SIGKDD Conference on Knowledge Discovery and Data Mining}{August 14--18,
2022}{Washington, DC, USA}
\acmBooktitle{Proceedings of the 28th ACM SIGKDD Conference on Knowledge
Discovery and Data Mining (KDD '22), August 14--18, 2022, Washington, DC, USA}
\acmPrice{15.00}
\acmDOI{10.1145/3534678.3539227}
\acmISBN{978-1-4503-9385-0/22/08}

\usepackage{amsthm}
\usepackage{algorithm}
\usepackage{algpseudocode}
\usepackage{amsmath}
\usepackage{bm}
\usepackage{color}
\usepackage{etex}
\usepackage{qtree}
\usepackage{graphicx}
\usepackage{multirow}
\usepackage{multicol}
\usepackage{subfigure}
\usepackage{url}
\usepackage{thmtools}

\usepackage{tabularx}
\usepackage{balance}    % balance the two columns of the paper in the last page
\usepackage{booktabs}   % professional tables. See http://cs.brown.edu/about/system/managed/latex/doc/booktabs.pdf
\usepackage{graphics}   % image files in figure (e.g., pdf, eps files)
\usepackage{url}        % formating a web url using \url{...}
\usepackage{pifont}     % adding special characters using \ding{...}. See http://willbenton.com/wb-images/pifont.pdf

\usepackage{xcolor}      % color text
\usepackage{xspace}
\usepackage{balance}

\newcommand{\ie}[0]{\textit{i.e.},\ }   % i.e., meaning "which is/means "
\newcommand{\eg}[0]{\textit{e.g.},\ }   % e.g., meaning "for example, "
\newcommand{\etc}[0]{\textit{etc.}\ }   % etc. meaning "and so on."

\usepackage[labelfont={bf,small},textfont={bf,small}]{caption}
\usepackage{pifont}% http://ctan.org/pkg/pifont
\newcommand{\cmark}{\ding{51}}%
\newcommand{\xmark}{\ding{55}}%

\usepackage{pgfplots}
\usepackage{ctable}

\newcommand{\eat}[1]{}

\begin{document}
%\title{A Multi-stage Anti-blur Network for Unsupervised Deformable Image Registration}

%\title{ERNet: An End-to-End Approach for Unsupervised Joint Brain Extraction and Registration}

%\title{ERNet: An End-to-End Approach for Unsupervised Joint Extraction and Registration in Neuroimaging Data}
%\title{ERNet: Unsupervised Collective Extraction and Registration Networks for Neuroimaging Data}

\title{ERNet: Unsupervised Collective Extraction and Registration in Neuroimaging Data}

\author{Yao Su}
\affiliation{%
\institution{Worcester Polytechnic Institute}
\city{Worcester}
\state{MA}
\country{USA}
\postcode{01609}}
\email{ysu6@wpi.edu}

\author{Zhentian Qian}
\affiliation{%
\institution{Worcester Polytechnic Institute}
\city{Worcester}
\state{MA}
\country{USA}
\postcode{01609}}
\email{zqian@wpi.edu}

\author{Lifang He}
\affiliation{%
\institution{Lehigh University}
\city{Bethlehem}
\state{PA}
\country{USA}
\postcode{18015}}
\email{lih319@lehigh.edu}

\author{Xiangnan Kong}
\affiliation{%
\institution{Worcester Polytechnic Institute}
\city{Worcester}
\state{MA}
\country{USA}
\postcode{01609}}
\email{xkong@wpi.edu}

\renewcommand{\shortauthors}{Yao Su et al.}

\begin{abstract}
Brain extraction and registration are important preprocessing steps in neuroimaging data analysis, where the goal is to extract the brain regions from MRI scans ({\ie} extraction step) 
%and align them into one coordinate system by transformation.
and align them with a target brain image ({\ie} registration step).
%Background
%Conventional methods often rely on visual inspection to filter out inaccurate extraction results before performing subsequent registration.
Conventional research mainly focuses on developing methods for the extraction and registration tasks separately under supervised settings. 
The performance of these methods highly depends on the amount of training samples and visual inspections performed by experts for error correction.
%
%Challenge
% The state-of-the-art paradigm based on multi-stage deformation and the improvement on registration accuracy unavoidably come with sharpness loss.
%However, conducting manual quality control in high-dimensional medical images (e.g., 3D MRI) is time-consuming, labor-intensive, and imprecise.
However, in many medical studies, collecting voxel-level labels and conducting manual quality control in high-dimensional neuroimages (\eg 3D MRI) are very expensive and time-consuming.
Moreover, brain extraction and registration are highly related tasks in neuroimaging data and should be solved collectively.
%
%However, current methods for multi-stage image registration are mainly focused
%Significance
%However, maintaining image quality such as sharpness during image registration is crucial to medical data analysis.
%In this paper, we propose an end-to-end unsupervised approach to accomplish both tasks simultaneously, called Extraction-Registration Network (ERNet).
In this paper, we study the problem of unsupervised collective extraction and registration in neuroimaging data.
%where no labeled image is required for extraction or registration.
%the goal is to extract brain regions from neuroimages and align them with a target image using only unlabeled data.
%
%We propose an end-to-end unsupervised approach to accomplish both tasks simultaneously, called Extraction-Registration Network (ERNet).
We propose a unified end-to-end framework, called ERNet (Extraction-Registration Network), to jointly optimize the extraction and registration tasks, allowing feedback between them.
%by learning from unlabeled images only.
%
%Research Gap & Solution
Specifically, we use a pair of multi-stage extraction and registration modules to learn the extraction mask and transformation, where the extraction network improves the extraction accuracy incrementally and the registration network successively warps the extracted image until it is well-aligned with the target image.
%Evidence
%We compare our work with the state-of-the-art extraction and registration methods, and demonstrate that our proposed ERNet achieves significantly better extraction performance on the publicly available datasets with comparable registration accuracy.
Experiment results on real-world datasets show that our proposed method can effectively improve the performance on extraction and registration tasks in neuroimaging data.

\end{abstract}
\keywords{brain extraction; skull stripping; registration; unsupervised; collective; multi-stage}

\ccsdesc[500]{Information systems~Data mining}
\ccsdesc[500]{Computing methodologies~Neural networks}
\ccsdesc[500]{Computing methodologies~Image segmentation}
\ccsdesc[500]{Computing methodologies~Matching}

% \ccsdesc[500]{Information systems~Data mining}
% \ccsdesc[500]{Computing methodologies~Neural networks}



\maketitle

% Introduction Section
%\vspace{-6pt}
\section{Introduction}
\label{sec:intro}

\textbf{Background.} Brain extraction (\emph{a.k.a.} skull stripping) and registration are preliminary but crucial steps in many neuroimaging studies.
%Examples of these studies include anatomical and functional analysis, radiotherapy, and surgical assistance.
Examples of these studies include anatomical and functional analysis~\cite{bai2017unsupervised,wang2017structural},
multi-modality fusion~\cite{cai2018deep}, diagnostic assistance~\cite{sun2009mining}.
The brain extraction step aims to remove the non-cerebral tissues such as the skull, dura, and scalp from the Magnetic Resonance Imaging (MRI) scan of a patient's head, while the registration step aims to align the extracted brain region with a template image of the standard brain. 
The extraction and registration steps are essential preprocessing procedures in many neuroimaging studies. 
For example, in anatomical and functional analysis, after removing and aligning the brain regions, the interference of non-neural tissues, imaging modalities, viewpoints can be eliminated, thus allowing precise quantification of changes in the shape, size, and position of anatomy and function. 
In brain atrophy diagnosis, a patient’s brain region across different pathological stages needs to be first extracted from raw brain MRI scans and then aligned with a standard template to counteract the non-diagnostic changes. These essential processing steps help doctors accurately monitor the alteration of brain volume. 
%-------------------------------------------------------
% \begin{figure}
%     \centering
%     \begin{minipage}[l]{\columnwidth}
%         \centering
%   	    \includegraphics[width=\textwidth]{fig/problem.pdf}
%         \vspace{-20pt}
%     \end{minipage}
% \caption{
% \textbf{Resource-efficient MTL. }
% Three tasks are being learned on a multi-task dataset. The goal is to train a model for all three tasks that can be deployed on mobile devices with limited computational resources.
% %\tian{remove tag "the author"...therefore, you could remove the emoji?}
% }
% \label{fig:problem}
% \vspace{-15pt}
% \end{figure}
%-------------------------------------------------------

\begin{figure*}[t]
  \centering
  \includegraphics[width=\linewidth]{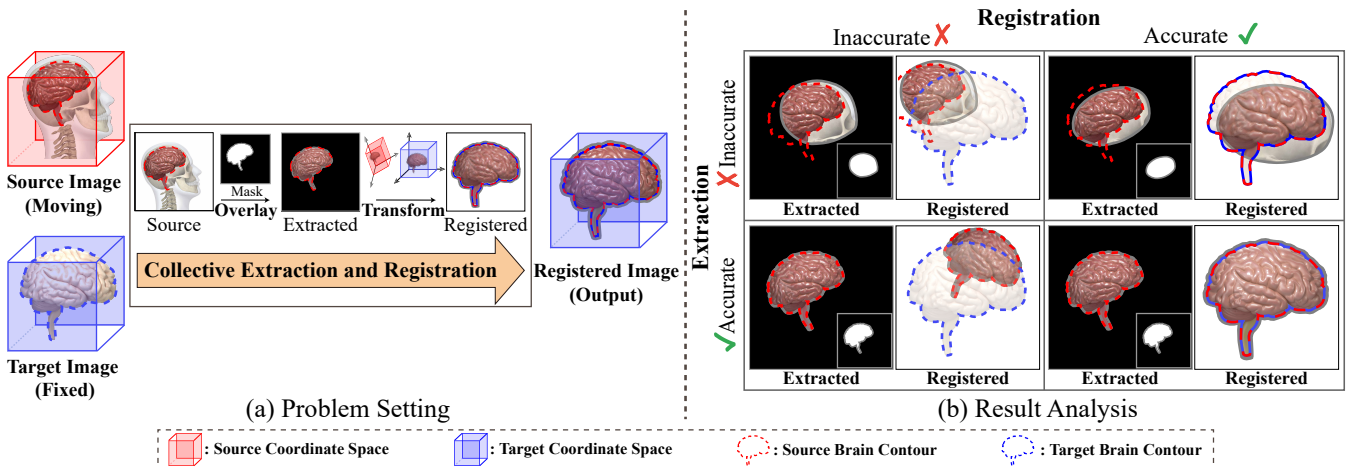}
  \vspace{-10pt}
  \caption{
  %An example of the anti-blur deformable image registration problem. 
The problem of unsupervised collective extraction and registration in neuroimaging data. (a) Given a raw scan of a patient's head (the source image) and a template image of standard brain region (the target image), the goal is to extract the brain region from the source image, and transform it to align with the target image. Neither the extraction label (\textit{i.e.}, the brain region in the source image) nor the registration label (\textit{i.e.}, the transformation required to align the source with the target) is available.
 Examples of different possible results are shown in (b).
  The bottom-right box is the ideal result: the correct region of the brain in the source is extracted and is well-aligned with the target.
  %The quadrant chart demonstrates the performance evaluation of the task.
  %Ideally, the brain region should be well-extracted while aligning with the target image, as shown in the bottom right of the chart.
  % achieve clearness and alignment, as shown in the bottom right of the registered image matrix.
  }
  \label{fig:intro}
  \vspace{-10pt}
\end{figure*}

% Image registration, as another essential step, aims to estimate the transformation between two or more images, and align them into one coordinate system.
% Within the same coordinate system, the impact of imaging modalities, viewpoint, and time can be eliminated, thus allowing precise quantification of changes in the shape, size, and position of anatomy and function.
% For example, a patient's brain MRI scans across different pathological stages can be fused together to counteract non-diagnostic movement, which helps doctors accurately monitor the growth of the tumor.
% Conventional methods for image registration often have drawbacks such as intensive computation and local minimum problems due to unique characteristics of medical images, \eg high dimensionality and heterogeneous pixel intensities.
% Recent studies tackle these problems by leveraging deep learning techniques, which significantly improve the registration speed while ensuring competitive accuracy.
% Nevertheless, these works typically rely on manual quality control to filter out inaccurate extraction results before performing subsequent registration.
% Conducting such visual inspection is not only time-consuming and labor-intensive, but also suffers from intra- and inter-rater variability.

\begin{figure}[t]
    \centering
    \subfigure[\textbf{Supervised  extraction
    \cite{kleesiek2016deep,lucena2019convolutional} + supervised registration \cite{sokooti2017nonrigid, dai2020dual} }]{
        \includegraphics[width=0.98\linewidth]{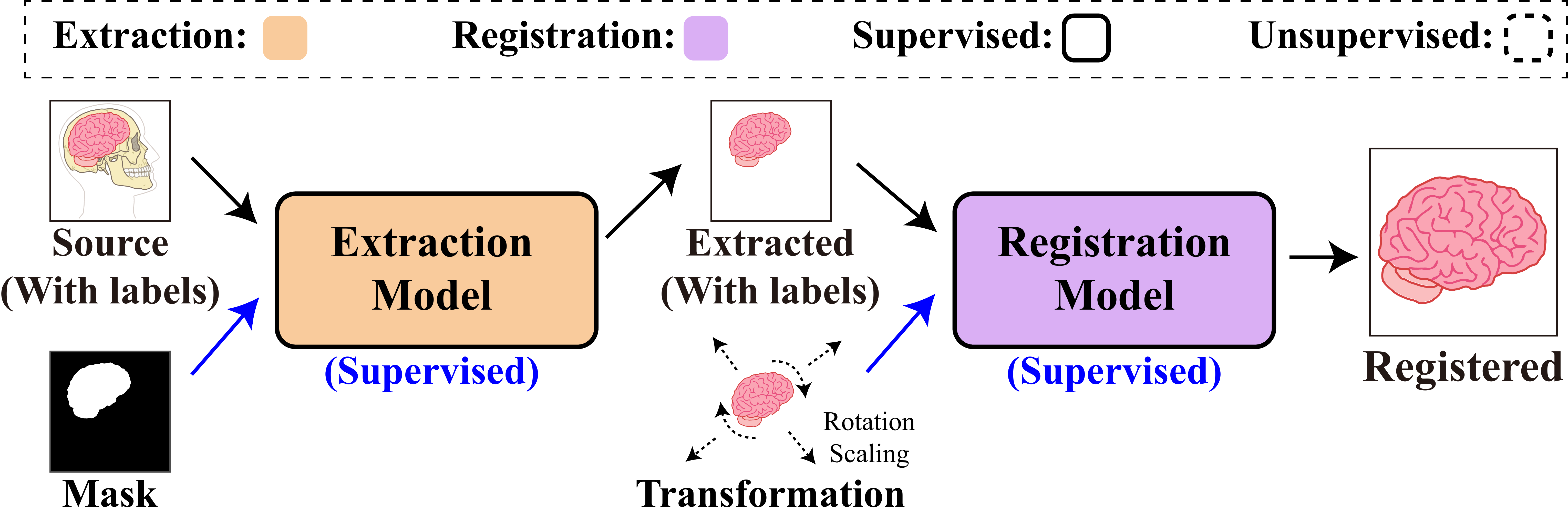}
        \label{fig:family 1}
    }

    \subfigure[\textbf{
        %Pipeline-based Extraction \cite{balakrishnan2018unsupervised} + Unsupervised Registration \cite{balakrishnan2018unsupervised}
        Unsupervised extraction \cite{smith2002fast,cox1996afni,shattuck2002brainsuite,segonne2004hybrid} + unsupervised registration \cite{balakrishnan2018unsupervised,zhao2019recursive}
        }]{
        \includegraphics[width=0.98\linewidth]{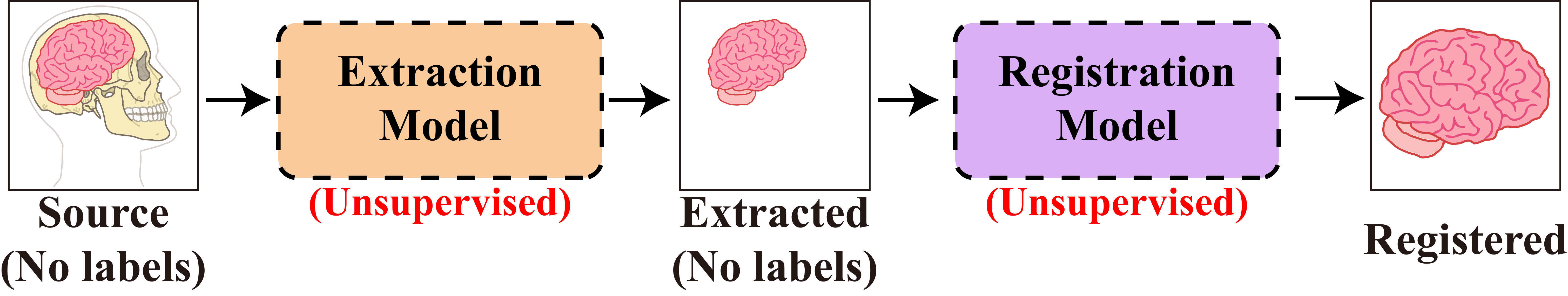}
        \label{fig:family 2}
    }
    \subfigure[\textbf{Unsupervised collective extraction and registration (ours)}]{
        \includegraphics[width=0.98\linewidth]{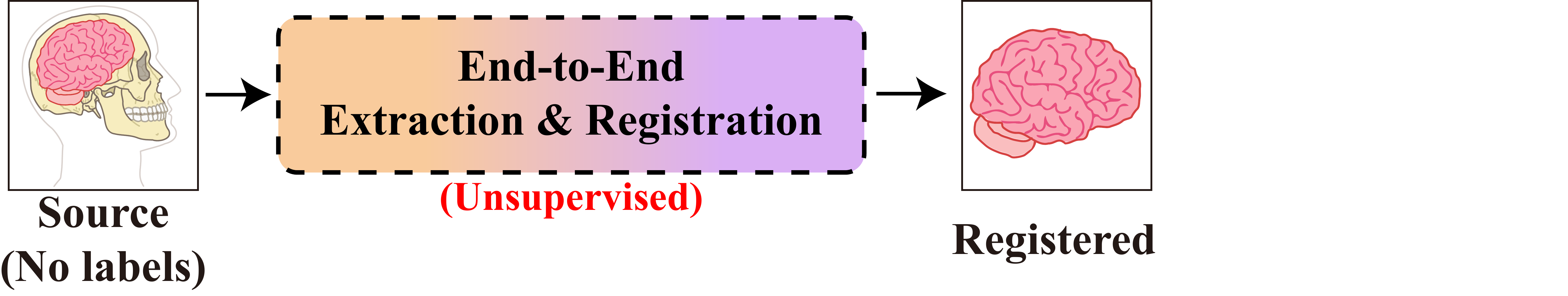}
        \label{fig:family 3}
        
    }
    \vspace{-10pt}
    \caption{
    %Comparing different solutions to the brain extraction and registration problem.
    Related works in brain extraction and registration.
    }
    \label{fig: family}
    \vspace{-20pt}
\end{figure}

\textbf{State-of-the-Art.} In the literature, brain extraction and registration problems have been extensively studied~\cite{kleesiek2016deep,lucena2019convolutional,sokooti2017nonrigid, dai2020dual}. Conventional approaches focus on developing methods for extraction~\cite{kleesiek2016deep,lucena2019convolutional} and registration~\cite{sokooti2017nonrigid, dai2020dual} separately under supervised settings, as shown in Figure~\ref{fig:family 1}.
However, in many medical studies, obtaining annotations of brain regions and transformations between images is often expensive as expertise, effort, and time are needed to produce precise labels, especially for high-dimensional neuroimages, \eg 3D MRI.
To address this limitation, recent works~\cite{smith2002fast,cox1996afni,shattuck2002brainsuite,segonne2004hybrid,balakrishnan2018unsupervised,zhao2019recursive} introduce a two-step approach for unsupervised extraction and registration by using automated brain extraction tools\cite{smith2002fast,cox1996afni,shattuck2002brainsuite,segonne2004hybrid} and unsupervised registration models~\cite{balakrishnan2018unsupervised,zhao2019recursive}, as shown in Figure~\ref{fig:family 2}.
Nevertheless, these works typically rely on manual quality control to correct inaccurate extraction results before performing subsequent registration.
Conducting such visual inspection is not only time-consuming and labor-intensive, but also suffers from intra- and inter-rater variability, thus limiting the efficiency and performance of both tasks.
More importantly, most existing methods still conduct extraction and registration separetely and neglect the potential relationship between these two tasks. 

\textbf{Problem Definition.} In this paper, we study the problem of unsupervised collective brain extraction and registration, as shown in Figure~\ref{fig:intro}(a).
Our goal is to capture the correlation of two tasks to boost their performance in an unsupervised setting. 
Specifically, the brain region needs to be extracted from the source image accurately and well-aligned to the target image without any labeled data.

\textbf{Challenges.} Despite its value and significance, the problem of unsupervised collective  extraction and registration has not been studied before and is very challenging due to its unique characteristics listed below:
% \begin{itemize}
%     \item \textit{Lack of labels in extraction:} Conventional learning-based extraction approaches are trained with a large number of training samples with ground truth labels. However, collecting voxel-level labels is very expensive and time-consuming in high-dimensional neuroimaging data.
%     \item \textit{Lack of labels in registration:} The ground truth transformation between the source and target images is difficult to obtain. Though there are unsupervised registration methods~\cite{balakrishnan2018unsupervised,zhao2019recursive} that optimize the transformation parameters by maximizing the similarity between images, these methods are only effective when the non-brain tissue of the source image is removed; otherwise, an erroneous transformation will be produced, rendering the registration invalid. Accordingly, obtaining the accurate transformation between the source and target images in an unsupervised setting remains largely unsolved.
%     \item \textit{Dependencies between extraction and registration:}
%     Conventional research mainly focuses on conducting extraction and registration tasks separately. However, these two tasks are highly correlated. Brain extraction has a decisive impact on the accuracy of the registration task, as shown in Figure~\ref{fig:intro}(b). At the same time, the registration task can help the extraction task to capture cerebral/non-cerebral information from the source and target images. Thereby, a holistic solution is desired to manage the interdependence between the two tasks. 
% \end{itemize}

\textbullet  \  
\textit{Lack of labels for extraction:} Conventional learning-based extraction approaches are trained with a large number of training samples with ground truth labels. However, collecting voxel-level labels is very expensive and time-consuming in high-dimensional neuroimaging data.

\textbullet  \  
\textit{Lack of labels for registration:} The ground truth transformation between source and target images is difficult to obtain. Though there are unsupervised registration methods~\cite{balakrishnan2018unsupervised,zhao2019recursive} that optimize the transformation parameters by maximizing the similarity between images, these methods are only effective when the non-brain tissue of the source image is removed; otherwise, an erroneous transformation will be produced, rendering the registration invalid. Thus, obtaining the accurate transformation between source and target images in an unsupervised setting remains largely unsolved.

\textbullet  \  
\textit{Dependencies between extraction and registration:} Conventional research mainly focuses on conducting extraction and registration tasks separately. However, these two tasks are highly correlated. Brain extraction has a decisive impact on the accuracy of the registration task, as shown in Figure~\ref{fig:intro}(b). Meanwhile,  registration task can help extraction task to capture cerebral/non-cerebral information from the source and target images. Thereby, a holistic solution is desired to manage the interdependence between the two tasks.

%---------------------

\textbf{Proposed Method.} To tackle the above issues, we propose a unified end-to-end framework, called ERNet (Extraction-Registration Network) for unsupervised collective brain extraction and registration.
Figure~\ref{fig: family} illustrates the comparison between our method and other state-of-the-art approaches.
Specifically, ERNet contains a pair of multi-stage extraction and registration modules, where the multi-stage extraction module progressively removes the non-brain tissue from the source image to produce an extracted brain image, and the multi-stage registration module incrementally aligns the extracted image to the target image.
These two modules help each other to boost extraction and registration performance simultaneously. 
The unalignable portion, \ie non-brain tissue,  revealed in the registration module guides the refinement process in the extraction module.
Meanwhile, the registration module benefits from accurate brain extraction generated in the extraction module. 
By bringing these two modules end-to-end, we achieve a joint optimization with no labels assistance. 

%\textbf{Contributions.} Our main contributions are summarized below:
%
%\textbullet  \  
%To the best of our knowledge, this is the first work to study the problem of unsupervised collective extraction and registration in neuroimaging data.
%
%\textbullet  \  
%We propose an end-to-end extraction and registration method that does not require any labeled brain masks and known transformations for training.
%
%\textbullet  \  
We design a new regularization term to smooth the predicted brain mask during the training, which improves the extraction accuracy to a certain extent.
%\textbullet  \  
Extensive experiments are performed on multiple public brain MRI datasets. The results indicate that our proposed method significantly outperforms state-of-the-art approaches in both extraction and registration accuracy.

\section{PRELIMINARIES} 
In this section, we first introduce related concept and notations, then define the unsupervised collective brain extraction and registration problem formally.

\subsection{Notations and Definitions}

\noindent \textbf{Definition 1 (Source and target images).}
Suppose we are given a training dataset $\mathcal{D} = \left\{\left(\mathbf{S}_i,\mathbf{T}_i\right)\right\}_{i=1}^{Z}$ that consists of $Z$ pairs of training samples. Each pair contains a source image $\mathbf{S}_{i} \in \mathbb{R}^{W \times H \times D}$ (\eg the raw MRI scan of a patient's head) and a target $\mathbf{T}_{i} \in \mathbb{R}^{W \times H \times D}$ (\eg a standard template of the brain region).
Here $W$, $H$ and $D$ denote the width, height and depth dimensions of the 3D images. 
For simplicity, we assume that the source and target images are resized to the same dimension, \ie $W \times H \times D$. 
Generally, in $\mathcal{D}$, the target images in different pairs can be different.
For example, in the cross-modality studies~\cite{cai2018deep}, we need to align the functional MRI (\ie source image) with the structural MRI (\ie target image) for each patient in the study (\ie an image pair in $\mathcal{D}$), where different patients will have different structural MRI images.
%\kong{can we add a short example of the application , ,when we need to use different target images for each pair? similar to the example we give below. such as cross modality analysis, aligning function to structure image.}
%\yao{Done.}
In many neuroimaging studies, however, all pairs in $\mathcal{D}$ can share a same target image, which is a special case of the dataset $\mathcal{D}$. 
For example, in brain network analysis, the functional MRI images (\ie source images) of all patients need to be aligned with a same template image (\ie target image), \eg MNI 152~\cite{sun2009mining}. 
%Focusing on neuroimaging applications, we consider the target image to be the same in this paper. 
For simplicity, in the following discussion, we omit the subscript $i$ of $\mathbf{S}_{i}$ and $\mathbf{T}_{i}$. 

\noindent \textbf{Definition 2 (Brain extraction mask).}
Brain extraction mask $\mathbf{M} \in \{0,1\}^{W \times H \times D}$ is a binary tensor of the same dimensions as the source image $\mathbf{S}$. 
$1$ in $\mathbf{M}$ corresponds to the cerebral tissues on $\mathbf{S}$ at the same location and $0$ otherwise.
The extracted image $\mathbf{E} = \mathbf{S} \circ \mathbf{M}$ is generated by applying the $\mathbf{M}$ on $\mathbf{S}$ via a element-wise product operator $\circ$.

\noindent \textbf{Definition 3 (Affine transformation and warped image).}
Without loss of generality, here we assume that the transformation in the registration task is affine-based. 
However, this work can be easily extended to other types of registration, \eg  nonlinear/deformable registration.
The affine transformation parameters $\mathbf{a} \in \mathbb{R}^{12} $ is a vector used to parameterized an affine transformation matrix $\mathbf{A} \in \mathbb{R}^{4 \times 4} $. The warped image $\mathbf{W} = \mathcal{T}\left(\mathbf{E},\mathbf{a}\right)$ is generated by applying the affine transformation on the extracted image $\mathbf{E}$, where $\mathcal{T}(\cdot, \cdot)$ is the affine transformation operator. The following relationship holds for $\mathbf{W}$ and $\mathbf{E}$ on the voxel level: 
\begin{equation}
\label{equ:voxel_value}
\mathbf{W}_{xyz} = \mathbf{E}_{x'y'z'},
\end{equation}
where the correspondences between coordinates $x,y,z$ and $x',y',z'$ are calculated based on the affine transformation matrix $\mathbf{A}$:
\begin{equation}
\begin{bmatrix}
x'\\
y'\\
z' \\
1
\end{bmatrix} = \mathbf{A}\begin{bmatrix}
x\\
y\\
z \\
1
\end{bmatrix} = \begin{bmatrix}
a_{1} & a_{2} & a_{3} & a_{4}\\
a_{5} & a_{6} & a_{7} & a_{8}\\
a_{9} & a_{10} & a_{11} & a_{12}\\
0 & 0 & 0 & 1
\end{bmatrix} \begin{bmatrix}
x\\
y\\
z \\
1
\end{bmatrix}.
\end{equation}
\begin{figure}[t]
  \centering
  \includegraphics[width=\linewidth]{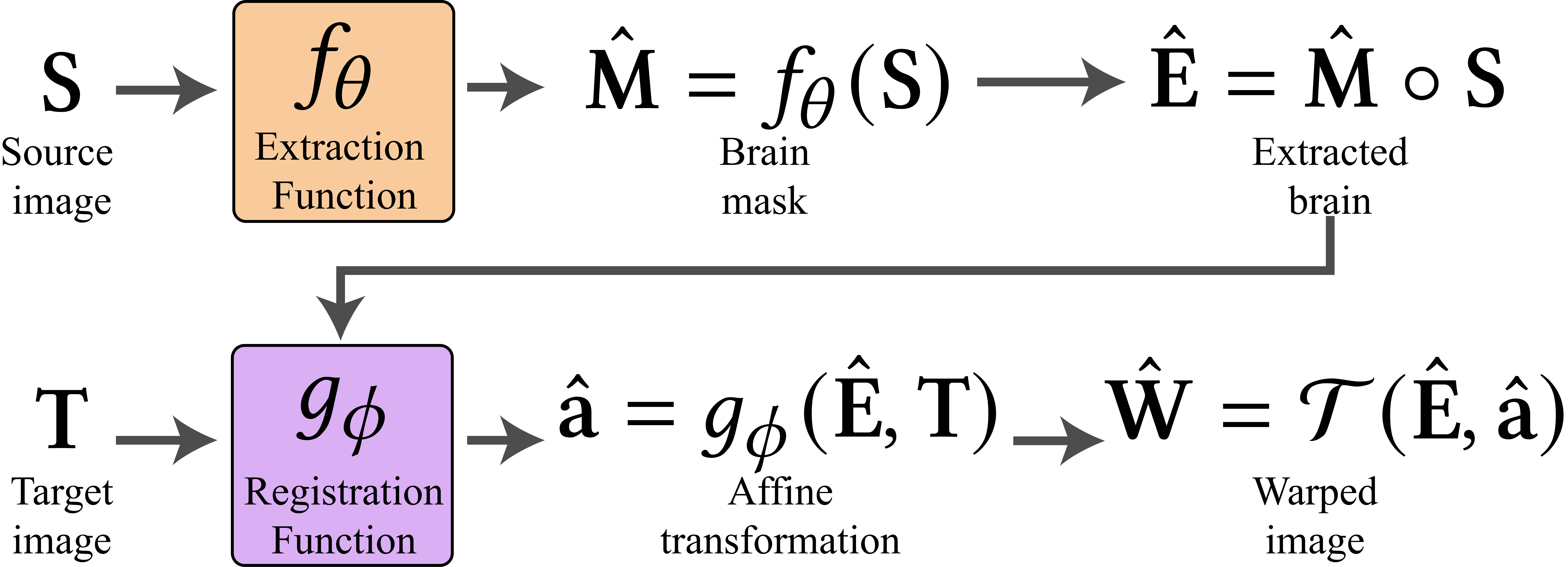}
  \vspace{-10pt}
  \caption{A demonstration of extraction and registration functions.}
  \label{fig:formulation}
  \vspace{-12pt}
\end{figure}
\subsection{Problem Formulation}

The goal of collective brain extraction and registration is to jointly learn the extraction function $f_{\theta}: \mathbb{R}^{W \times H \times D} \rightarrow \mathbb{R}^{W \times H \times D}$ and the registration function $g_{\phi}: \mathbb{R}^{W \times H \times D}\times \mathbb{R}^{W \times H \times D} \rightarrow \mathbb{R}^{12}$, as shown in Figure~\ref{fig:formulation}.
Specifically, the extraction function $f_{\theta}(\cdot)$ takes the source image $\mathbf{S}$ as input to predicts a brain extraction mask $\hat{\mathbf{M}} = f_{\theta}(\mathbf{S})$. Then, the registration function $g_{\phi}(\cdot, \cdot)$ takes the extracted brain image $\hat{\mathbf{E}} = \hat{\mathbf{M}} \circ \mathbf{S}$ and the target image $\mathbf{T}$ to predict the affine transformation parameter $\hat{\mathbf{a}} = g_{\phi}(\hat{\mathbf{E}},\mathbf{T})$. Finally, the warped image is $\hat{\mathbf{W}} = \mathcal{T}(\hat{\mathbf{E}},\hat{\mathbf{a}})$. The optimal parameter $\theta^*$ and $\phi^*$ can be found by solving the following optimization problem:
\begin{equation}
\begin{split}
\label{equ:goal_training}
\theta^{*},\phi^{*} &=\underset{\theta,  \phi}{\arg \min } \hspace{-3pt} \sum_{\left(\mathbf{S},  \mathbf{T}\right)\in \mathcal{D}}\left[ \mathcal{L} \left( \hat{\mathbf{W}}, \mathbf{T}  \right)\right] \\
& = \underset{\theta, \phi}{\arg \min } \hspace{-3pt}  \sum_{\left(\mathbf{S}, \mathbf{T}\right)\in \mathcal{D}}\left[ \mathcal{L} \left( \mathcal{T} \left( f_{\theta}(\mathbf{S}) \circ \mathbf{S},\hspace{5pt} g_{\phi} \left(  f_{\theta}(\mathbf{S}) \circ \mathbf{S}, \mathbf{T}  \right)\right),   \mathbf{T}    \right)\right],
\end{split}
\end{equation}
where the image pair $(\mathbf{S},\mathbf{T})$ is sampled from the training dataset $\mathcal{D}$, and $\mathcal{L}(\cdot, \cdot)$ is image dissimilarity criteria, \eg mean square error.

%\yao{This formula seems too complicated. I am not sure whether we should just write it as $\theta^{*},\phi^{*} =\underset{\theta, \hspace{1pt} \phi}{\arg \min } \hspace{1pt} \sum_{\left(\mathbf{S}, \hspace{1pt} \mathbf{T}\right)\in \mathcal{D}}\left[ \mathcal{L} \left( \hat{\mathbf{W}}, \mathbf{T}  \right)\right]  $}

% Specifically, this function takes a test pair of source image $\mathbf{S}$ and target image $\mathbf{T}$ and predict an tuple of extraction mask and affine transformation parameter $(\mathbf{M}, \mathbf{a}) = f_{\theta}(\mathbf{S},\mathbf{T})$. 
% The predicted brain extraction mask $\mathbf{M}$ is applied on the $\mathbf{S}$ to remove the non-cerebral tissues and generate the extracted image $\mathbf{E}$. We further apply the affine transformation parameterized by $\mathbf{a}$ on $\mathbf{E}$ to align extracted cerebral tissues with $\mathbf{T}$. The optimal parameter $\theta^*$ of function $f_{\theta}(\cdot, \cdot)$ is found by solving the following optimization problem:
% \begin{equation}
% \label{equ:goal_training}
% \theta^{*} =  \underset{\theta}{\arg \min } \hspace{1pt} \sum_{\left(\mathbf{S}, \hspace{1pt} \mathbf{T}\right)\in \mathcal{D}}\left[\mathcal{L}\left(\mathcal{T}\left(\mathbf{S}\circ \mathbf{M}, \mathbf{a}\right),\hspace{1pt} \mathbf{T}\right)\right],
% \end{equation}
% where $(\mathbf{M}, \mathbf{a}) = f_{\theta}(\mathbf{S},\mathbf{T})$, the image pair $(\mathbf{S},\mathbf{T})$ is sampled from the training dataset $\mathcal{D}$, and $\mathcal{L}(\cdot, \cdot)$ is image dissimilarity criteria, \eg mean square error. 

To the best of our knowledge, this work is the first endeavour to find an optimal solution to the problem of unsupervised collective brain image extraction and registration in an end-to-end neural network. Our approach excludes the necessity of labeling the brain extraction masks and transformation between images of the training dataset, as opposed to other supervised methods~\cite{kleesiek2016deep,lucena2019convolutional,sokooti2017nonrigid, dai2020dual}.

% %-----------------------------------------------
% % Method Section

\definecolor{myorange}{RGB}{249, 203, 156}
\definecolor{myviolet}{RGB}{218, 175, 244}

\begin{figure*}[t]
  \centering
  \includegraphics[width=0.98\linewidth]{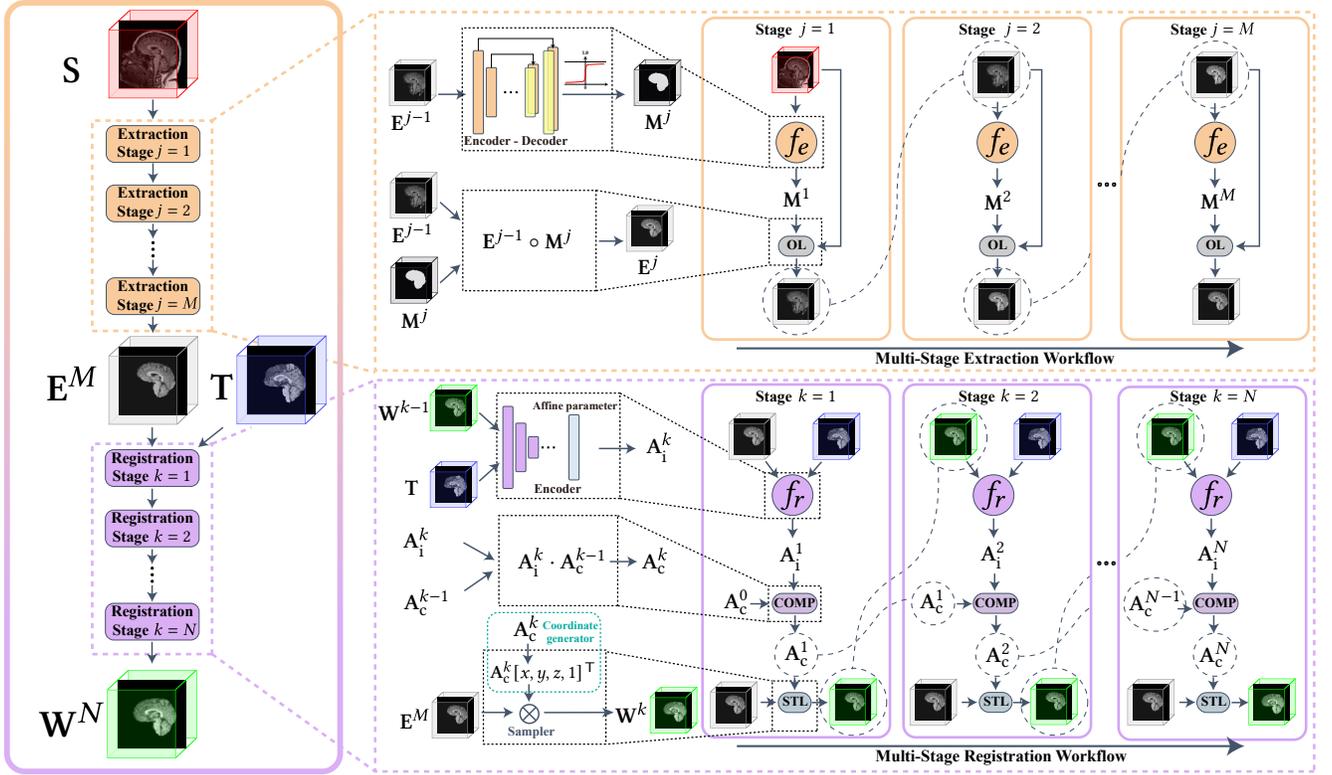}
    \vspace{-10pt}
  \caption{An overview of ERNet for collective brain extration and registration. \textcolor{myorange}{Multi-stage extraction module:} In each extraction stage $j$, the \emph{ Extraction Network} $f_{e}$ predicts the current mask $\mathbf{M}^{j}$ from previous extracted image $\mathbf{E}^{j-1}$; the \emph{Overlay Layer} (OL) performs an element-wise product between the previous extracted image $\mathbf{E}^{j-1}$ and the current mask $\mathbf{M}^{j}$ to generate the current extracted image $\mathbf{E}^{j}$. The final extracted image is $\mathbf{E}^{M}$. \textcolor{myviolet}{Multi-stage registration module:} In each registration stage $k$, the \emph{Registration Network} $f_{r}$ predicts the current affine transformation $\mathbf{A}_\text{i}^{k}$ between the previous warped image $\mathbf{W}^{k-1}$ and target image $\mathbf{T}$; the \emph{Composition Layer} (COMP) fuses current affine transformation $\mathbf{A}_\text{i}^{k}$ and previous combined affine transformation $\mathbf{A}_\text{c}^{k-1}$ to generate the updated combined affine transformation $\mathbf{A}_\text{c}^{k}$; the \emph{Spatial Transformation Layer} (STL) performs the transformation $\mathbf{A}_\text{c}^{k}$ on the final extracted image $\mathbf{E}^{M}$ to produce the warped image $\mathbf{W}^{k}$. The final output image is $\mathbf{W}^{N}$.
    }
  \label{fig:network}
  \vspace{-10pt}
\end{figure*}

%An overview of ABN for multi-stage deformable image registration.In each stage $k$, the \emph{Short-Term Registration Network} $f_{S}$ predicts the current deformation field $\phi_{S}^{(k)}$ between the previous warped image $\mathbf{I}_{w}^{(k-1)}$ and target image $\mathbf{I}_{t}$; \emph{Long-Term Memory Network} $f_{L}$ fuses current deformation field $\phi_{S}^{(k)}$ and previous combined deformation field $\phi_{L}^{(k-1)}$ to generate the updated combined deformation field $\phi_{L}^{(k)}$; \emph{Spatial Transformation Layer} (ST Layer) performs the nonlinear transformation on the source image $\mathbf{I}_{s}$ using $\phi_{L}^{(k)}$ to produce the warped image $\mathbf{A}_{i}^{(1)}$. The final warped registered image is $\mathbf{A}_{c}^{(1)} \cdot \big(\mathbf{I}_{e}^{(j-1)} \circ \mathbf{M}^{(j)}\big)$.
\section{Our Approach}
\label{sec:method}
\noindent\textbf{Overview.} Figure~\ref{fig:network} presents an overview of the proposed ERNet framework for the unsupervised collective brain extraction problem. Our method is a multi-stage deep neural network consisting of two main modules: 1) \emph{Multi-Stage Extraction Module} takes the raw source image $\mathbf{S}$ as input, and gradually produces the extracted brain image $\mathbf{E}^{M}$ after $M$ stages of extraction; 2) \emph{Multi-Stage Registration Module} takes the extracted brain image $\mathbf{E}^{M}$ and the target image $\mathbf{T}$ as inputs, and incrementally aligns $\mathbf{E}^{M}$ with $\mathbf{T}$ through $N$ stages of registration. The final output is the warped image $\mathbf{W}^{N}$. The whole framework is trained using backpropagation in an end-to-end unsupervised fashion, allowing feedback and collaboration between modules.
Next we introduce the details of each module and the training process.
\subsection{Multi-Stage Extraction Module}
\label{sec: Extration Net}
This module proposes to solve the extraction task in a multi-stage fashion to obtain high extraction accuracy. It contains $M$ extraction stages with each stage $j$ consisting of two main components: 1) \emph{Extraction Network} takes the previous extracted brain image $\mathbf{E}^{j-1}$ as input, and generates a current brain mask $\mathbf{M}^{j}$; 2) \emph{Overlay Layer} takes the previous extracted brain image $\mathbf{E}^{j-1}$ and the current extraction mask $\mathbf{M}^{j}$ as inputs, and generates an updated extracted image $\mathbf{
E}^{j}$. The output of this module is the extracted brain image $\mathbf{E}^{M}$ at the final stage $j = M$.

\subsubsection{Extraction Network: $f_{e}$}
\label{sec: fe}
The extraction network $f_{e}(\cdot)$ serves to gradually remove the non-cerebral tissues in the source image $\mathbf{S}$ so that only cerebral tissues would remain on the extracted image at the final stage. At each stage $j$, based on the extracted brain image $\mathbf{E}^{j-1}$ from the previous stage, it would produce a current extraction mask $\mathbf{M}^{j}$ to remove the non-cerebral tissues believed to be still on $\mathbf{E}^{j-1}$. Specifically, we adopt 3D U-Net \cite{ronneberger2015u} as the base network to learn $f_{e}(\cdot)$, which is the state-of-the-art architecture widely used in image registration and semantic segmentation. The output of the U-Net would go through a Heaviside step function to obtain the binary mask $\mathbf{M}^{j}$ when performing inference:
\begin{equation}
 H(x) = \begin{cases}
        1, & \text{if } x > 0, \\
        0, & \text{otherwise}.
    \end{cases}
\end{equation}
Note that the derivative of the Heaviside step function does not exist for $x =0$ and is a constant $0$ for $x\neq 0$. For the gradient to successfully backpropagate, we use a Sigmoid function with a large slope parameter $\gamma$ to approximate the Heaviside step function when performing training:
\begin{equation}
S(x) = \frac{1}{1 +e^{-\gamma x}}
\label{eq:sigmoid}
\end{equation}

$f_{e}(\cdot)$ follows a shared-weight design, which means that $f_{e}(\cdot)$ is repetitively applied across stages with the same parameters. It can be formalized as:
\begin{equation}
\mathbf{M}^{j}=f_{e}\left(\mathbf{E}^{j-1}\right),
\end{equation}
where $\mathbf{M}^{j}$ is the outputted brain mask of the $j$-th stage for $j = [1, \cdots, M]$ and $\mathbf{E}^{0} = \mathbf{S}$.

\subsubsection{Overlay Layer: $OL$}
The overlay layer would remove the non-cerebral tissues remaining in the image by applying the current brain mask $\mathbf{M}^{j}$ to the previous extracted image $\mathbf{E}^{j-1}$. The updated extracted image is:
\begin{equation*}
    \mathbf{E}^{j} = \mathbf{E}^{j-1}\circ \mathbf{M}^{j}
\end{equation*}
where $\circ$ is the element-wise product operator.

\subsection{Multi-Stage Registration Module}
Similar to the extraction module discussed in Section \ref{sec: Extration Net}, we implement a multi-stage solution to address the registration task. This module consists of $N$ cascaded stages with each stage $k$ containing three main components: 1) \emph{Registration Network} takes the previous warped image $\mathbf{W}^{k-1}$ and the target image $\mathbf{T}$ as inputs, and generates the current affine transformation $\mathbf{A}_\text{i}^{k}$; 2) \emph{Composition Layer} takes the previous combined affine transformation $\mathbf{A}_\text{c}^{k-1}$ and the current affine transformation $\mathbf{A}_\text{i}^{k}$ as input, and generates an updated combined affine transformation $\mathbf{A}_\text{c}^{k}$; and 3) \emph{Spatial Transformation Layer} transforms the extracted brain image $\mathbf{E}^{M}$ using $\mathbf{A}_\text{c}^{k}$ to produce the warped image $\mathbf{W}^{k}$. 
%Note that in the $k$-th cascaded stage, the extracted brain image $\mathbf{E}^{M}$ and the warped image $\mathbf{W}^{k-1}$ are jointly utilized to minimize the sharpness loss during the successive affine transformation operations.  
The output of this module is the warpped brain image $\mathbf{W}^{N}$ at the final stage.

\subsubsection{Registration Network: $f_{r}$}
The registration network $f_{r}(\cdot, \cdot)$ is designed to gradually transform the extracted brain image to maximize its similarity with the target image. At each stage $k$, it predicts a current affine transformation $\mathbf{A}_\text{i}^{k}$ relying only on the previous warped image $\mathbf{W}^{k-1}$ and the target image $\mathbf{T}$.
% An important reason to want a short-term transformation is to be able to control the transformation process more locally and more accurately, avoiding the under-transformation or over-transformation of the previous warped image. 
Following a similar approach to the extraction network $f_e$ in Section \ref{sec: fe}, we adopt a 3D CNN based encoder to learn $f_{r}(\cdot, \cdot)$ and a shared weight design to utilize $f_{r}(\cdot, \cdot)$ repetitively across stages with the same parameters. 
%  It is possible to estimate the long-term overall performance athlete or estimates in the short term for each stage of preparation or participation in competitions 
% In order to avoid under/over deformation of the source image to the extent which the source image may become identical to the target image 
% This structure is comprised of an encoder-decoder with skip connections to integrate hierarchical information between low-level and high-level features. 
It can be formalized as:
\begin{equation}
\mathbf{A}_\text{i}^{k}=f_{r}\left(\mathbf{W}^{k-1}, \mathbf{T}\right),
\end{equation}
where $\mathbf{A}_\text{i}^{k}$ is the output affine transformation of the $k$-th stage for $k = [1, \cdots, N]$ and $\mathbf{W}^{0} = \mathbf{E}^{M}$.

\subsubsection{Composition Layer: $\text{COMP}$}
In each stage $k$, after obtaining the current affine transformation $\mathbf{A}_\text{i}^{k}$ from $f_{r}(\cdot, \cdot)$, we would combine all previous transformation:
\begin{equation}
\mathbf{A}_\text{c}^{k}=\mathbf{A}_\text{i}^{k} \cdot \mathbf{A}_\text{c}^{k-1},
\end{equation}
where $\cdot$ is matrix product. When $k=1$, the initial affine transformation $\mathbf{A}_\text{c}^{0}$ is set to be an identity matrix representing no displacement. This layer serves as a bridge between the registration network and the spatial transformation layer. 
% As such, it enables the utilization of both the extracted brain image $\mathbf{I}_\text{e}^{(m)}$ and the warped image $\mathbf{I}_\text{w}^{(k-1)}$ to preserve the image sharpness for better details.
As such, it enables the transformation to be directly applied to the final extracted brain image $\mathbf{E}^{M}$ to avoid image sharpness loss caused by multiple interpolations.

% This network structure allows it to exhibit dynamic constitutive behavior as it contains the interconnect between different stages of deformations. Similar to $f_{R}(\cdot)$, we adopt an U-Net based CNN to learn $\text{COMP}(\cdot)$ with a weight sharing across each stage. 

% Other different network architectures can be chosen, if they prove to have better performance. 

\subsubsection{Spatial Transformation Layer}
An important step towards image registration is to reconstruct the warped image $\mathbf{W}^{k}$ from the extracted brain image $\mathbf{E}^{M}$ by affine transformation operator. Based on the combined transformation $\mathbf{A}_\text{c}^{k}$, we introduce a spatial transformation layer to resample the voxels into a uniform grid on the extracted image to acquire the warped image through $\mathbf{W}^{k} = \mathcal{T}(\mathbf{E}^{M}, \mathbf{A}_\text{c}^{k})$. According to the definition of 
affine transformation operator in Eq.~(\ref{equ:voxel_value}), we have
\begin{equation}
     \mathbf{W}_{xyz}^{k} = \mathbf{E}_{x'y'z'}^{M} \hspace{5pt},
     \label{equ:voxel_value_k}
\end{equation}
where $[x', y', z', 1]^\top = \mathbf{A}_\text{c}^{k}[x, y, z, 1]^\top $.

% resampling technique
% enable interpolation of linear operators

To ensure the success of gradient propagation in this process, we use a differentiable transformation based on trilinear interpolation introduced by \cite{jaderberg2015spatial}. That is, 
\begin{equation}
\begin{split}
 \mathbf{W}_{xyz}^{k}= \sum_{o=1}^{W} \sum_{p=1}^{H}  \sum_{q=1}^{D} \mathbf{E}_{opq}^{M} &\cdot  
\max(0,1-|x' -o|) \\ 
 \cdot  \max (0,1-|y' - p|) & \cdot \max (0,1-|z' - q|).
\end{split}
\end{equation}
Notice that Eq.~(\ref{equ:voxel_value_k}) always performs transformation on the extracted image $\mathbf{E}^{M}$ instead of the previous warped images. Therefore, only one interpolation is required to produce the final warped brain image $\mathbf{W}^{N}$, which better preserves the sharpness of $\mathbf{W}^{N}$.

\subsection{Unsupervised End-to-End Training}
\label{section:end-to-end training}

We train our ERNet model in an unsupervised setting by minimizing the following objective function
\begin{equation}
\label{eq:loss}
\underset{\mathbf{M}^{1}, \cdots, ~\mathbf{M}^{M}, ~\mathbf{A}_\text{c}^{N}}{\min} \mathcal{L}_{\operatorname{sim}} \left(\mathbf{W}^{N}, \mathbf{T}\right)+ \sum_{j=1}^{M}\lambda \mathcal{R}(\mathbf{M}^{j}),
\end{equation}
where $\mathbf{W}^{N} = \mathcal{T}(\mathbf{E}^{M}, \mathbf{A}_\text{c}^{N})$ and $\mathcal{L}_{\operatorname{sim}}(\cdot, \cdot)$ is a loss function measuring the similarity between the final warped image $\mathbf{W}^{N}$ and the target image $\mathbf{T}$. Here we use the popular negative local cross-correlation loss, which is robust to voxel intensity variations often found across scans and datasets \cite{balakrishnan2018unsupervised}. $\mathcal{R}(\cdot)$ is the regularization term for brain masks, and $\lambda$ is a regularization parameter. Since the brain region is one connected entity, we would like our predicted brain masks to have the same properties across all stages. To put it more formally, if we view the brain mask as a 3D tensor with 6-connectivity, we would like it to have exactly one connected component. Though it is possible to count the number of connected components in a mask, such practice would be time-consuming ($O(W\times H\times D)$ by BFS algorithm) and not differentiable. For the purpose of both effective and efficient estimation, we use the $\ell_2$-norm of the first-order derivative of $\mathbf{M}^{j}$ as the regularization term:
%\begin{equation}
%\mathcal{R}(\mathbf{M}^{j})=\sum\|\nabla %\mathbf{M}^{j}\|^{2} = .
%\end{equation}

\begin{equation}
\mathcal{R}(\mathbf{M}^{j})=\sum_{x=1}^{W} \sum_{y=1}^{H}  \sum_{z=1}^{D}\|\nabla \mathbf{M}_{xyz}^{j}\|^{2}.
\end{equation}

This regularization term measures edge strength, \ie the likelihood of a voxel to be an edge voxel. By minimizing the regularization term, we can suppress the occurrence of edges, which in term suppress additional connected components. Specifically, we approximate the first-order derivative by measuring differences between neighboring voxels.
For $\nabla\mathbf{M}_{xyz}^{j}=(\frac{\partial \mathbf{M}_{xyz}^{j}}{\partial x}, \frac{\partial \mathbf{M}_{xyz}^{j}}{\partial y}, \frac{\partial \mathbf{M}_{xyz}^{j}}{\partial z})$, we have $\frac{\partial \mathbf{M}_{xyz}^{j}}{\partial x} \approx \mathbf{M}_{(x+1)yz}^{j} - \mathbf{M}_{xyz}^{j} $. The approximation of $\frac{\partial \mathbf{M}_{xyz}^{j}}{\partial y}$ and $\frac{\partial \mathbf{M}_{xyz}^{j}}{\partial z}$ follows.

%This regularization term would suppress the occurrence of edges, which in term suppress additional connected components. Specifically, we approximate the spatial gradients by measuring differences between neighboring voxels.
%For $\nabla\mathbf{M}_{xyz}^{j}=\left(\frac{\partial \mathbf{M}_{xyz}^{j}}{\partial x}, \frac{\partial \mathbf{M}_{xyz}^{j}}{\partial y}, \frac{\partial \mathbf{M}_{xyz}^{j}}{\partial z}\right)$, we approximate 

Benefiting from the differentiability of each component of this design, our model can be cooperatively and progressively optimized across each stage in an end-to-end manner. Such a training scheme allows us to find a  joint optimal solution to the collective brain extraction and registration task.

\section{Experiments}

\subsection{Datasets} \label{section:Dataset}
%In order to evaluate our proposed ABN method, we conduct experiments on both natural and medical image registration tasks, by using the following datasets.
%In order to evaluate the performance of our method on image registration tasks, we conduct experiments on three different datasets, one with natural images (2D) and two with medical images (3D).

We evaluate the effectiveness of our proposed method on three different real-world 3D brain MRI datasets: 1) \emph{LPBA40}~\cite{shattuck2008construction} consists of 40 raw T1-weighted 3D brain MRI scans along with their brain masks. It also provides the corresponding segmentation ground truth of 56 anatomical structures; 2) \emph{CC-359}~\cite{souza2018open} consists of 359 raw T1-weighted 3D brain MRI scans and the corresponding brain masks. It also contains the labeled white matter as the ground truth; 3) \emph{IBSR}~\cite{rohlfing2011image} provides 18 raw T1-weighted 3D brain MRI scans along with the corresponding manually segmentation results. Due to the small sample size, We use this dataset only to test the model trained on CC359.
The brain mask and anatomical segmentations are used to evaluate the accuracy of extraction and registration, respectively. Datasets are splited into training, validation, and test sets, respectively. For more details, please refer to Appendix~\ref{sec:appendix}.

\subsection{Compared Methods}
\label{section:Compared Methods}

%We compare our approach to seven state-of-the-art methods, as shown in Table~\ref{tab:methods}.

We compare our ERNet with several representative brain extraction and registration methods, as shown in Table~\ref{tab:methods}. To the best of our knowledge, there is no existing solution that can perform the brain extraction and registration simultaneously under an unsupervised setting. Therefore, we designed two-stage pipelines for comparison using the following brain extraction and registration methods. 

% \noindent\textbullet\ \textit{Brain Extraction Tool (BET)} \cite{smith2002fast}: This is a skull stripping method included in FMRIB's Software Library (FSL). It uses a deformable approach to fit the brain surface by applying locally adaptive set models.

\noindent\textbullet\ \textit{Brain Extraction Tool (BET)} \cite{smith2002fast}: This is a skull stripping method included in FSL package. It uses a deformable approach to fit the brain surface by applying locally adaptive set models.

% \noindent\textbullet\ \textit{Brain Extraction Tool$^*$ (BET$^*$)} \cite{smith2002fast}: This is an optimized version of BET, which separates the non-brain and brain regions by an intensity-based estimation at a threshold of 0.7.

% \noindent\textbullet\ \textit{3dSkullStrip} \cite{cox1996afni}: This is a modified version of BET that is included in the Analysis of Functional Neuro Images (AFNI) package. It performs skull stripping based on the expansion paradigm of the spherical surface.
\noindent\textbullet\ \textit{3dSkullStrip} \cite{cox1996afni}: This is a modified version of BET that is included in the AFNI package. It performs skull stripping based on the expansion paradigm of the spherical surface.

\noindent\textbullet\ \textit{Brain Surface Extractor (BSE)} \cite{shattuck2002brainsuite}: It extracts the brain region based on morphological operations and edge detection, which employs anisotropic diffusion filtering and a Marr Hildreth edge detector for brain boundary identification.

% \noindent\textbullet\ \textit{FMRIB's Linear Image Registration Tool (FLIRT)} \cite{jenkinson2001global}: This is a fully automated linear (affine) brain image registration tool that is part of the FMRIB's Software Library (FSL).
\noindent\textbullet\ \textit{FMRIB's Linear Image Registration Tool (FLIRT)} \cite{jenkinson2001global}: This is a fully automated affine brain image registration tool in FSL package.

% \noindent\textbullet\ \textit{Advanced Normalization Tools (ANTs)} \cite{avants2009advanced}: It is popularly considered a state-of-the-art medical image registration toolkit, which contains various transformation models and similarity metrics for image registration. Here we utilize affine transformation model and cross-correlation metric for registration.

\noindent\textbullet\ \textit{Advanced Normalization Tools (ANTs)} \cite{avants2009advanced}: It is considered a state-of-the-art medical image registration toolkit. Here we utilize affine transformation model and cross-correlation metric for registration.

\noindent\textbullet\ \textit{VoxelMorph (VM)} \cite{balakrishnan2018unsupervised}: This unsupervised, deformable image registration method employs a neural network to predict the nonlinear transformation between images.

\noindent\textbullet\ \textit{Cascaded Registration Networks (CRN)} \cite{zhao2019recursive}:
It is an unsupervised multi-stage image registration method. In different stages, the source image is repeatedly deformed to align with a target image.

\noindent\textbullet\ \textit{ERNet}: This is our proposed model which consists of both extraction and registration modules in an end-to-end manner.

\noindent\textbullet\ \textit{ERNet w/o Ext}: This is a variant of ERNet where we remove the extraction modules. Here it is a registration method only.

\begin{table}[t]
    \centering
    \caption{Summary of compared methods.}
    \label{tab:methods}
    \vspace{-10pt}
    \resizebox{\linewidth}{!}{
    \begin{tabular}{lcccc}
    \toprule
    \textbf{Methods}& \textbf{Extraction}& \textbf{Registration}& \textbf{Collaborative}& \textbf{Deep learning}\\
    \midrule
    BET { \cite{smith2002fast}} & \cmark  & \xmark & \xmark & \xmark \\
    % BET$^*$ { \cite{smith2002fast}} & \cmark  & \xmark & \cmark & \xmark\\
    3dSkullStrip \cite{cox1996afni}& \cmark  & \xmark & \xmark & \xmark\\
    BSE \cite{shattuck2002brainsuite}& \cmark  & \xmark & \xmark & \xmark\\
    \midrule
    FLIRT \cite{jenkinson2001global} & \xmark  & \cmark & \xmark & \xmark\\
    ANTs \cite{avants2009advanced}& \xmark  & \cmark & \xmark & \xmark\\
    VM \cite{balakrishnan2018unsupervised}& \xmark  & \cmark & \xmark & \cmark\\
    CRN \cite{zhao2019recursive}& \xmark  & \cmark & \xmark & \cmark \\
    \midrule
    ERNet w/o Ext& \xmark  & \cmark & \xmark & \cmark\\
    ERNet (ours)& \cmark  & \cmark & \cmark & \cmark\\
    \bottomrule
    \end{tabular}
    }
    \vspace{-15pt}
\end{table}
%%%%% No time %%%%%%%%%%%
\begin{table*}[t]
    \centering
    \caption{Results for brain extraction and registration in different datasets. The results are reported as performance(mean $\pm$ std ) of extraction and registration of each compared method. The running time is reported as the average processing time for each image in its corresponding task.
    “$\uparrow$” point out “the larger the better” and “$\downarrow$” point out “the smaller the better”.}
    \label{tab:main res}
    \vspace{-10pt}
    \resizebox{1.0\linewidth}{!}{
    \begin{tabular}{lcccccccrc}
    \toprule
    \multicolumn{2}{c}{\multirow{2}{*}{Methods}}                & \multicolumn{6}{c}{Datasets}                                                        & \multicolumn{2}{c}{\multirow{2}{*}{Running Time}} \\
    \cmidrule(lr){3-8}
    
    \multicolumn{2}{c}{}                                        & \multicolumn{2}{c}{LPBA40} & \multicolumn{2}{c}{CC359} & \multicolumn{2}{c}{IBSR}  & \multicolumn{2}{c}{}                              \\
    
    \cmidrule(lr){1-2}\cmidrule(lr){3-4}\cmidrule(lr){5-6}\cmidrule(lr){7-8}\cmidrule(lr){9-10}
    
    \multirow{2}{*}{Extraction} & \multirow{2}{*}{Registration} & Extraction  & Registration & Extraction & Registration & Extraction & Registration & Extraction             & Registration             \\

                                &                               & Dice$_{\mathrm{ext}}$  $\uparrow$        & Dice$_{\mathrm{reg}}$ $\uparrow$         & Dice$_{\mathrm{ext}}$ $\uparrow$       & Dice$_{\mathrm{reg}}$ $\uparrow$         & Dice$_{\mathrm{ext}}$ $\uparrow$       & Dice$_{\mathrm{reg}}$ $\uparrow$         & Sec $\downarrow$                    & Sec $\downarrow$ \\

    \midrule
    % BET \cite{smith2002fast} & FLIRT \cite{jenkinson2001global} & 0.865 $\pm$ 0.106 & 0.455 $\pm$ 0.300 & 0.751 $\pm$ 0.140 & 0.709 $\pm$ 0.096 & 0.904 $\pm$ 0.054 & 0.790 $\pm$ 0.034 & 2.86 (CPU) & 4.06 (CPU)\\
    
    BET \cite{smith2002fast} & FLIRT \cite{jenkinson2001global} & 0.935 $\pm$ 0.028 & 0.606 $\pm$ 0.026 & 0.811 $\pm$ 0.087 & 0.747 $\pm$ 0.060 & 0.911 $\pm$ 0.038 & 0.798 $\pm$ 0.010 & 2.45 (CPU) & 4.57 (CPU)\\
    
    3dSkullStrip \cite{cox1996afni} & FLIRT \cite{jenkinson2001global}
    & 0.902 $\pm$ 0.032 & 0.594 $\pm$ 0.018 & 0.849 $\pm$ 0.037 & 0.790 $\pm$ 0.034 & 0.869 $\pm$ 0.039 & 0.787 $\pm$ 0.020 & 178.56 (CPU) & 4.64 (CPU)\\
    
    BSE \cite{shattuck2002brainsuite} & FLIRT \cite{jenkinson2001global}
    & 0.938 $\pm$ 0.022 & 0.614 $\pm$ 0.010 & 0.846 $\pm$ 0.112 & 0.801 $\pm$ 0.021 & 0.873 $\pm$ 0.064 & 0.798 $\pm$ 0.015 & 4.75 (CPU) & 4.35 (CPU) \\
    
    \midrule
    
    % BET \cite{smith2002fast} & ANTs \cite{avants2009advanced} & 0.865 $\pm$ 0.106 & 0.617 $\pm$ 0.013 & 0.751 $\pm$ 0.140 & 0.737 $\pm$ 0.094 & 0.904 $\pm$ 0.054 & 0.796 $\pm$ 0.026 & 2.86 (CPU) & 2.96 (CPU)\\
    
    BET \cite{smith2002fast} & ANTs \cite{avants2009advanced} & 0.935 $\pm$ 0.028 & 0.609 $\pm$ 0.025 & 0.811 $\pm$ 0.087 & 0.764 $\pm$ 0.053 & 0.911 $\pm$ 0.038 & 0.796 $\pm$ 0.014 & 2.45 (CPU)& 2.76 (CPU)\\
    
    3dSkullStrip \cite{cox1996afni} & ANTs \cite{avants2009advanced} & 0.902 $\pm$ 0.032 & 0.616 $\pm$ 0.016 & 0.849 $\pm$ 0.037 & 0.807 $\pm$ 0.027 & 0.869 $\pm$ 0.039 & 0.794 $\pm$ 0.017 & 178.56 (CPU) & 2.89 (CPU)\\
    
    BSE \cite{shattuck2002brainsuite} & ANTs \cite{avants2009advanced} & 0.938 $\pm$ 0.022 & 0.616 $\pm$ 0.013 & 0.846 $\pm$ 0.112 & 0.796 $\pm$ 0.027 & 0.873 $\pm$ 0.064 & 0.797 $\pm$ 0.017 & 4.75 (CPU) & 2.52 (CPU)\\
    
    \midrule
    
    BET \cite{smith2002fast} & VM \cite{balakrishnan2018unsupervised}
    & 0.935 $\pm$ 0.028 & 0.488 $\pm$ 0.092 & 0.811 $\pm$ 0.087 & 0.811 $\pm$ 0.015 & 0.911 $\pm$ 0.038 & 0.792 $\pm$ 0.010 & 2.45 (CPU) & 0.02 (GPU)\\
    
    3dSkullStrip \cite{cox1996afni} & VM \cite{balakrishnan2018unsupervised}
    & 0.902 $\pm$ 0.032 & 0.479 $\pm$ 0.094 & 0.849 $\pm$ 0.037 & 0.809 $\pm$ 0.018 & 0.869 $\pm$ 0.039 & 0.785 $\pm$ 0.014 & 178.56 (CPU) & 0.02 (GPU) \\
    
    BSE \cite{shattuck2002brainsuite} & VM \cite{balakrishnan2018unsupervised}
    & 0.938 $\pm$ 0.022 & 0.512 $\pm$ 0.065 & 0.846 $\pm$ 0.112 & 0.810 $\pm$ 0.017 & 0.873 $\pm$ 0.064 & 0.794 $\pm$ 0.011 & 4.75 (CPU) & 0.02 (GPU)\\
    
    \midrule
    
    BET \cite{smith2002fast} & CRN \cite{zhao2019recursive}
    & 0.935 $\pm$ 0.028 & 0.556 $\pm$ 0.046 & 0.811 $\pm$ 0.087 & 0.815 $\pm$ 0.008 & 0.911 $\pm$ 0.038 & 0.799 $\pm$ 0.017 & 2.45 (CPU) & 0.10 (GPU) \\
    
    3dSkullStrip \cite{cox1996afni} & CRN \cite{zhao2019recursive}
    & 0.902 $\pm$ 0.032 & 0.528 $\pm$ 0.056 & 0.849 $\pm$ 0.037 & 0.813 $\pm$ 0.009 & 0.869 $\pm$ 0.039 & 0.796 $\pm$ 0.014 & 178.56 (CPU) & 0.10 (GPU)\\
    
    BSE \cite{shattuck2002brainsuite} & CRN \cite{zhao2019recursive}
    & 0.938 $\pm$ 0.022 & 0.547 $\pm$ 0.071 & 0.846 $\pm$ 0.112 & 0.812 $\pm$ 0.011 & 0.873 $\pm$ 0.064 & 0.799 $\pm$ 0.011 & 4.75 (CPU) & 0.10 (GPU)\\
    
    \midrule
    
    BET \cite{smith2002fast} & \textbf{ERNet} (w/o Ext)
    & 0.935 $\pm$ 0.028 & 0.616 $\pm$ 0.021 & 0.811 $\pm$ 0.087 & 0.804 $\pm$ 0.017 & 0.911 $\pm$ 0.038 & 0.798 $\pm$ 0.011 & 2.45 (CPU) & 0.04 (GPU)\\
    
    3dSkullStrip \cite{cox1996afni} & \textbf{ERNet} (w/o Ext)
    & 0.902 $\pm$ 0.032 & 0.606 $\pm$ 0.006 & 0.849 $\pm$ 0.037 & 0.815 $\pm$ 0.007 & 0.869 $\pm$ 0.039 & 0.791 $\pm$ 0.014 & 178.56 (CPU) & 0.04 (GPU)\\
    
    BSE \cite{shattuck2002brainsuite} & \textbf{ERNet} (w/o Ext)
    & 0.938 $\pm$ 0.022 & 0.613 $\pm$ 0.012 & 0.846 $\pm$ 0.112 & 0.807 $\pm$ 0.018 & 0.873 $\pm$ 0.064 & 0.795 $\pm$ 0.013 & 4.75 (CPU) & 0.04 (GPU)\\
    
    \midrule
    
    \multicolumn{2}{c}{\textbf{ERNet (ours)}} & \textbf {0.946 $\pm$ 0.009} & \textbf{0.626 $\pm$ 0.008} & \textbf{0.938 $\pm$ 0.008} & \textbf{0.818 $\pm$ 0.006} & \textbf{0.916 $\pm$ 0.013} & \textbf{0.800 $\pm$ 0.011} & \multicolumn{2}{c}{\textbf{0.11 (GPU)}}\\
    
    \bottomrule             
    \end{tabular}}
    \vspace{-10pt}
\end{table*}

\subsection{Evaluation Metrics}
\label{section:Evaluation Metrics}
%We use two metrics to assess the registered image quality: registration accuracy and image sharpness, which are detailed as follows. 
% Due to the differences between datasets used in the two tasks, separate evaluation metrics are used to assess the registration accuracy. 
% The 3D brain MRI datasets include the segmentation of anatomical structures that can be used to measure registration performance. However, the 2D face image dataset only contains different pairs of source images and target images. Thus their registration accuracy can only be evaluated by the similarity between the final warped images and the target images. Therefore, we use corresponding evaluation metrics for each task. 
% We introduce each corresponding evaluation metric in the following subsection.
Our defined problem aims to identify the brain region within the source image and align the extracted cerebral tissues to the target image simultaneously. Thus, we evaluate the accuracy of extraction and registration to show the performance of our proposed method and compared methods as follows:

\subsubsection{Extraction Performance.}
The brain MRI datasets contain the brain mask ground truth, which is the label of brain tissue in the source image. To evaluate the extraction accuracy, we measure the volume overlap of brain masks by Dice score, which can be formulated as:
%\begin{equation}
%\text{Dice}_{\mathrm{ext}}=2 \cdot \frac{|\mathbf{M} \cap %\widetilde{\mathbf{M}}|}{|\mathbf{M}|+|\widetilde{\mathbf{M}}|},
%\end{equation}
\begin{equation}
\text{Dice}_{\mathrm{ext}}=2 \cdot \frac{|\mathbf{\hat{M}} \cap \mathbf{M}|}{|\mathbf{\hat{M}}|+|\mathbf{M}|},
\end{equation}
where $\mathbf{\hat{M}}$ is the predicted brain mask and $\mathbf{M}$ denotes the corresponding ground truth. If $\mathbf{\hat{M}}$ represents accurate extraction, we expect the non-zero regions in $\mathbf{\hat{M}}$ and $\mathbf{M}$ to overlap well.

\subsubsection{Registration Performance.}
We evaluate the registration accuracy by measuring the volume overlap of anatomical segmentations, which are the location labels of different tissues in the brain MRI image. If two images are well aligned, then their corresponding anatomical structures should overlap with each other. Likewise, the Dice score can evaluate the performance of registration, as follows:
\begin{equation}
\text{Dice}_{\mathrm{reg}}=2 \cdot \frac{|\mathbf{G}_\text{w} \cap \mathbf{G}_\text{t}|}{|\mathbf{G}_\text{w}|+|\mathbf{G}_\text{t}|}
\end{equation}
where $\mathbf{G}_\text{w}=\mathcal{T}\left(\mathbf{G}_\text{s},\mathbf{\hat{a}}\right)$.
% \begin{equation}
% \mathbf{S}_\text{w}=\mathcal{T}\left(\mathbf{S}_\text{s}\circ \mathbf{M},\mathbf{a}\right).
% \end{equation}
$\mathbf{G}_\text{s}$, $\mathbf{G}_\text{t}$ and $\mathbf{G}_\text{w}$ are anatomical structural segmentation of the source image, target image and warped image, respectively. $\mathbf{\hat{a}}$ is the predicted affine transformation parameters. A dice score of 1 means that the corresponding structures are well aligned after registration, a score of 0 indicates that there is no overlap. If the image contains multiple labeled anatomical structures, the final score is the average of the dice score of the each structure.

\subsection{Experimental Results}
We compare our ERNet with the baseline models regarding extraction and registration accuracy. For each task, we quantify the performance by its corresponding dice score. We also report the running time for each method to complete both tasks. Across all these metrics, we show that ERNet not only consistently achieves better extraction and registration performance than other alternatives, but is also more time-efficient.

% \subsubsection{Experiment Setting.}
% We split the datasets into training and test sets as described in the Datasets section. 
% For 2D face registration, models are trained with batch size of 16 for 5$k$ epochs. 
% For 3D brain MRI registration, we reduce the batch size to 1 to address GPU memory limitation and models were trained for 1$k$ epochs. 
% For both tasks, models are optimized using Adam optimizer~\cite{kingma2014adam} with a learning rate of $1 \times 10^{-4}$. 
% The ratios of regularization in Eq.~\eqref{eq:loss} is set to $\lambda=10$.
% For more details, please refer to supplementary material. The source code is available at \url{https://github.com/anonymous3214/ABN}.
% % Implementation details and code are available in the supplementary material.
% % All models were implemented using PyTorch with the code available at 

\subsubsection{Experiment Setting.}
% We split the datasets into training sets and test sets as described in Section~\ref{section:Dataset}. 
% Note that the IBSR dataset is used for testing only. We train the deep neural networks using the machine learning framework PyTorch.
% Each training batch consists of one pair of images to address GPU memory limitation and models are optimized using Adam optimizer~\cite{kingma2014adam} with a learning rate of $1 \times 10^{-6}$. 
% We leverage image augmentation technique during training to expand the datasets, where a small random transformation with translation, rotation, and scaling is applied to images. For more details, please refer to supplementary material. 
% The source code is available at 
We split the datasets into training, validation, and test sets as described in Appendix~\ref{sec:appendix}. 
Note that the IBSR dataset is used for test only. 
The training set is used to learn model parameters and the validation set is used to evaluate the performance of hyperparameter settings (e.g., the number of stages or smoothing regularization term). 
We use the test set only once to report the final evaluation results for each model.
%Each training batch consists of one pair of images to address GPU memory limitation and models are optimized using Adam optimizer~\cite{kingma2014adam} with a learning rate of $1 \times 10^{-6}$. 
% We leverage the image augmentation technique during training to expand the datasets, where a small random transformation with translation, rotation, and scaling is applied to images.
%For more details, please refer to supplementary material. 
%The source code is available at 
%\url{https://github.com/ERNetERNet/ERNet}.

% Implementation details and code are available in the supplementary material.
% All models were implemented using PyTorch with the code available at 

% described in Section \ref{section:Compared Methods}.
\begin{figure}[t]
  \centering
  \includegraphics[width=\linewidth]{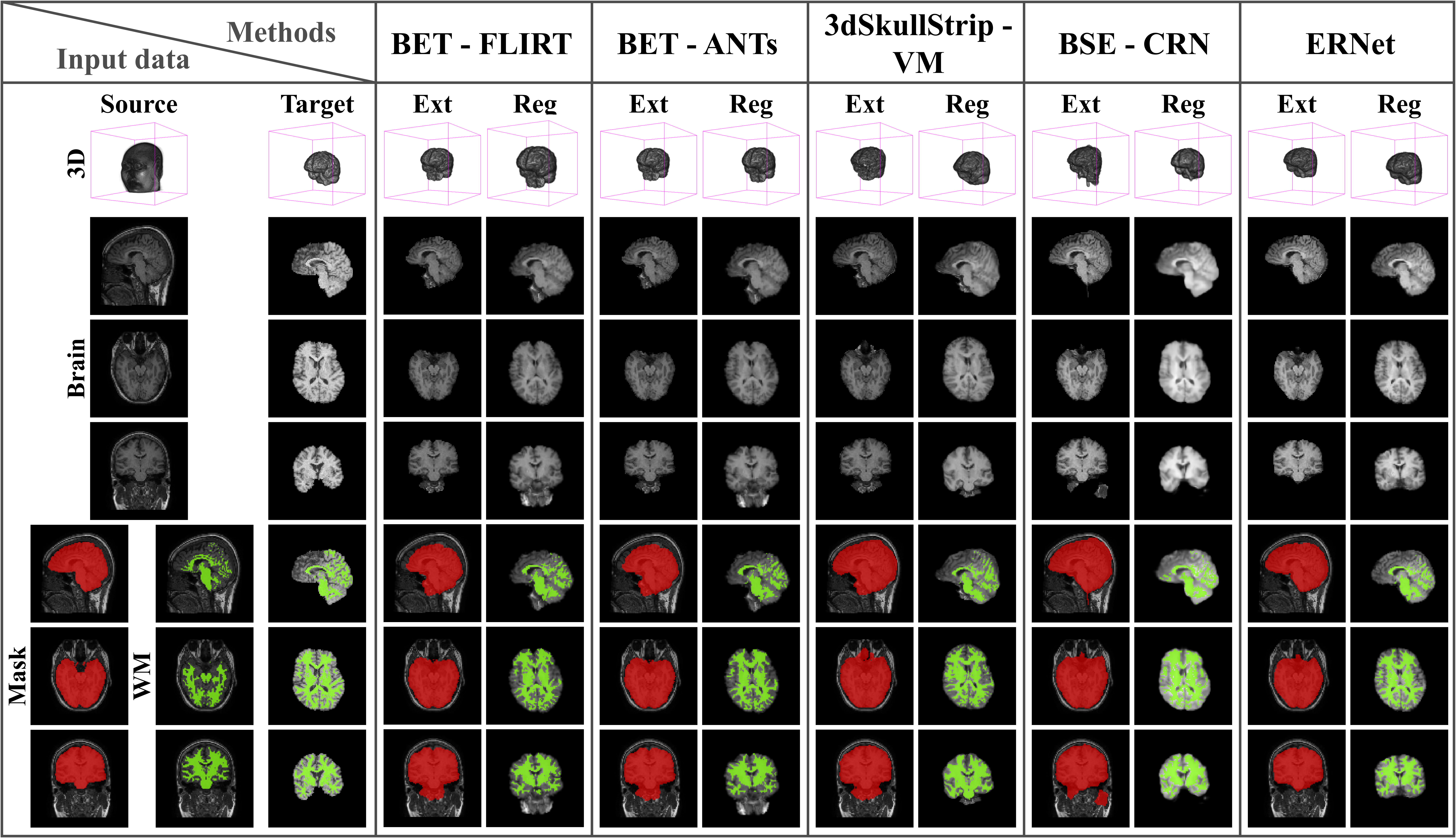}
  \vspace{-15pt}
  \caption{Visual comparisons for brain extraction and registration task. We render a 3D visualization of the image and display middle slice in three different planes: sagittal, axial and coronal. The right side contains the source and target images and their corresponding ground truth labels. We show extraction and registration results of each method and its corresponding predictive labels used for performance evaluations. To evaluate the brain extraction task, a predicted brain \textbf{Mask} {\color{red}(red)} should coincide as much as possible with the ground truth brain \textbf{Mask} {\color{red}(red)} of the source image. Likewise, in the brain registration task, a warped \textbf{W}hite \textbf{M}atter {\color{
 green}(green)} should well-overlap with the \textbf{W}hite \textbf{M}atter {\color{
 green}(green)} of the target image.}
  \label{fig:main result}
  \vspace{-20pt}
\end{figure}
\begin{figure}[t]
  \centering
  \includegraphics[width=\linewidth]{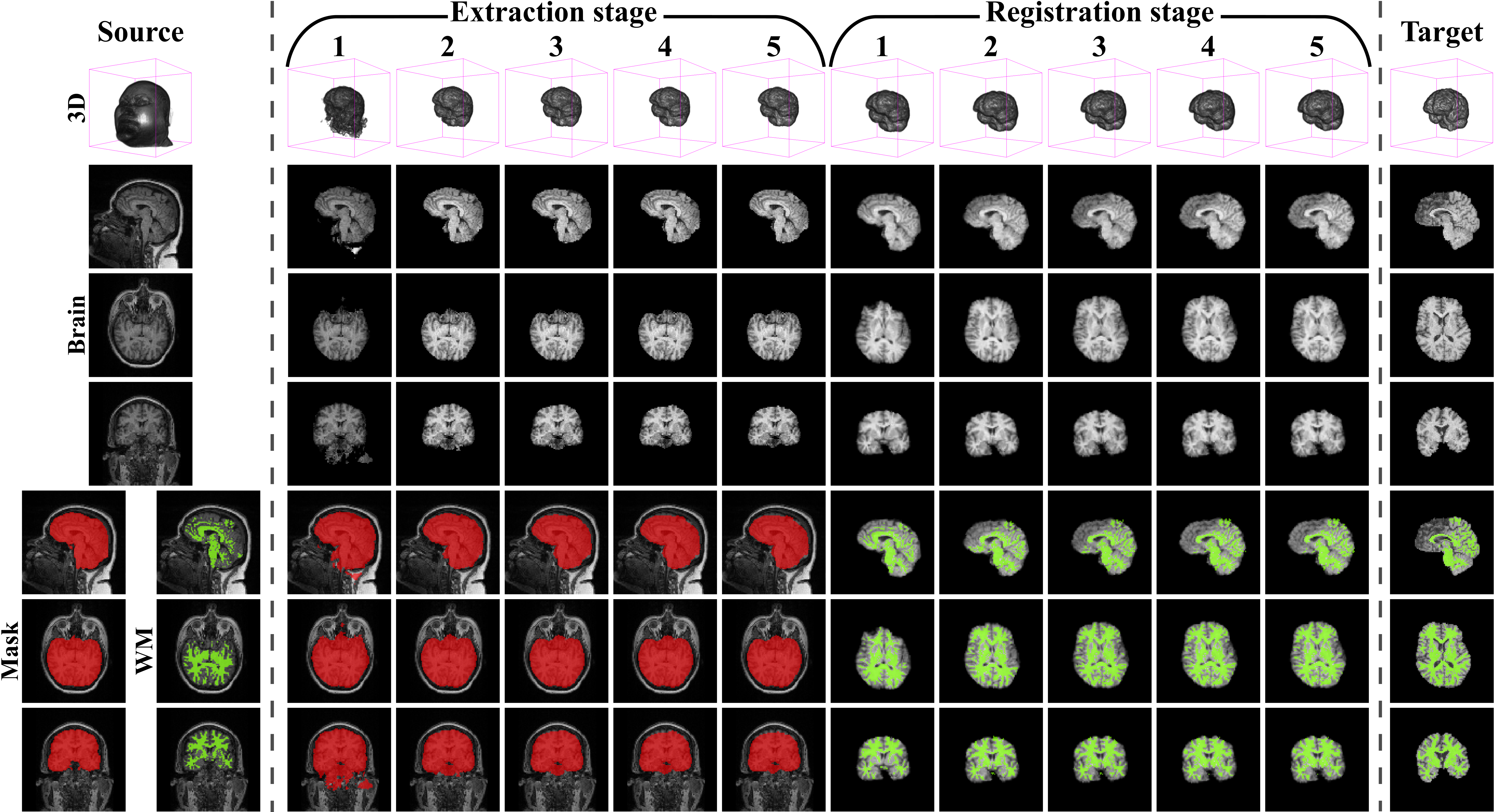}
  \vspace{-15pt}
  \caption{A demonstration of the multi-stage extraction and registration process. The example shows the extraction and registration result at each stage using an ERNet with 5-stage extraction and 5-stage registration.}
  \label{fig:stage result}
  \vspace{-20pt}
\end{figure}
%\input{tab_2d}
%\input{fig_res_2d}
% \vspace{-5}
\subsubsection{Extraction and Registration Results.}
Table~\ref{tab:main res} summarized the results of designed two-stage pipelines and proposed ERNet in both extraction and registration tasks.
Based on the global competition in three datasets, ERNet outperforms the existing methods in all metrics.
For the extraction task, the performance of ERNet is superior to that of the other compared methods, especially on the CC359 dataset.
Specifically, we observed a gain in extraction dice score up to $10.5\%$ compared to the best extraction method 3dSkullStrip.
Besides, ERNet is more robust in the extraction task than other alternatives, as it performs consistently well and obtains the smallest standard deviation across all datasets.

When observing registration performance, once again, ERNet outperforms all other methods across all datasets.
Most notably, we find that the registration result of almost every method is bounded by the result of its corresponding extraction method.
% , except for one datapoint outlier that occurred when testing the combination of BET$^{*}$ and CRN in the CC359 dataset. Even in the outlier case, the registration result is only slightly better than the extraction result.
This proves that the accuracy of extraction has a significant impact on the quality of the subsequent registration task.
ERNet captures this property to deliver an improved result via end-to-end collective learning.
In addition, the overall performance of ERNet w/o Ext, the variant of ERNet, is slightly worse than that of ERNet, indicating that the extraction network in ERNet is crucial and beneficial to improve both extraction and registration accuracy.  

\subsubsection{Running Efficiency.}
To evaluate the efficiency of ERNet, we compare its running time with other baselines. The run time is measured on the same machine with a Intel$^{\circledR}$ Xeon$^{\circledR}$ E5-2667 v4 CPU and an NVIDIA Tesla V100 GPU.
As shown in Table~\ref{tab:main res}, ERNet is roughly 20 to 200 times faster than existing methods.
This is because ERNet can perform both extraction and registration tasks end-to-end on the same device efficiently.
In contrast, other alternatives handle these two tasks separately using two-step approaches, which is time-consuming.
No GPU implementations of BET, 3dSkullStrip, BSE, FLIRT and ANTs are made available~\cite{smith2002fast,cox1996afni,shattuck2002brainsuite,jenkinson2001global,avants2009advanced}.

\subsubsection{Qualitative Analysis.}
Figure~\ref{fig:main result} shows visualized brain extraction and registration results of our ERNet compared with other two-step approaches on the LPBA40 test set.
Upon inspection, we can see that extracted image of ERNet is more accurate than those of BET, 3dSkullStrip, and BSE.
The brain mask predicted by ERNet overlaps best with the ground truth mask of the source image, while the masks predicted by other extraction methods contain obvious non-brain tissues.
In terms of registration results, ERNet also clearly outperforms other compared methods. 
The final registered image of ERNet is more similar to the target image than that of alternatives. 
Most notably, inaccurate extraction results with non-brain tissue also appear in the following registration results and hamper the final performance. 
This supports our claim that a failed extraction can propagate to the following registration task, rendering an irreversible error.
Furthermore, we demonstrate the intermediate extraction and registration results of ERNet in Figure~\ref{fig:stage result} using a test sample of CC359 dataset.
It is clear that brain tissue is progressively extracted from the source image with the help of a multi-stage design. Likewise, the extracted image is transformed multiple times to align with the target image incrementally.

%%%%% No time %%%%%%%%%%%
\begin{table}[t]
    \centering
    \caption{Influence of number of stages on extraction and registration performance. “$\uparrow$” point out “the larger the better”.}
    \label{tab: stage res}
    \vspace{-10pt}
    \resizebox{0.85\linewidth}{!}{
    \begin{tabular}{cccc}
    \toprule
    
    \multicolumn{2}{c}{Number of Stages} & Extraction & Registration \\
    
    \cmidrule(lr){1-2}
    
    Extraction       & Registration      & Dice$_{\mathrm{ext}}$ $\uparrow$       & Dice$_{\mathrm{reg}}$ $\uparrow$         \\

    \midrule
    0 & 0 & 0.216 $\pm$ 0.018 & 0.252 $\pm$ 0.158\\
    
    0 & 1 & 0.216 $\pm$ 0.018 & 0.269 $\pm$ 0.101\\ 
    
    0 & 5 & 0.216 $\pm$ 0.018 & 0.264 $\pm$ 0.103\\ 
    
    1 & 0 & 0.040 $\pm$ 0.026 & 0.007 $\pm$ 0.007\\ 
    
    1 & 1 & 0.902 $\pm$ 0.010 & 0.566 $\pm$ 0.040\\
    
    1 & 5 & 0.927 $\pm$ 0.007 & 0.604 $\pm$ 0.017\\
    
    5 & 0 & 0.095 $\pm$ 0.040 & 0.024 $\pm$ 0.016\\ 
    
    5 & 1 & 0.919 $\pm$ 0.010 & 0.550 $\pm$ 0.033\\
    
     \textbf{5} &  \textbf{5} &  \textbf{0.946 $\pm$ 0.009} &  \textbf{0.626 $\pm$ 0.008}\\
    
    \bottomrule
    \end{tabular}}
    \vspace{-10pt}
\end{table}
    
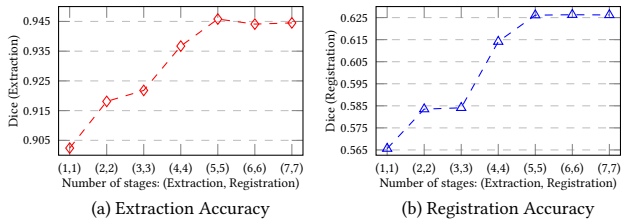
\begin{figure}[t]
\begin{tikzpicture}
\begin{axis}[
    title = {(a) Extraction Accuracy},
    outer sep=0,
    width=0.57\columnwidth,
    height=0.42\columnwidth,
    xlabel={Number of stages: (Extraction, Registration)},
x label style={at={(axis description cs:0.5,0.17)},anchor=north},
y label style={at={(axis description cs:0.254,.5)},anchor=south},
    % xlabel near ticks,
    % ylabel near ticks,
    ylabel={Dice (Extraction)},
    xmin=-0.3, xmax=6.3,
    ymin=0.9, ymax=0.950,
    xtick={0,1,2,3,4,5, 6},
    ytick={0.905,0.915,0.925,0.935,0.945},
    yticklabels={0.905,0.915,0.925,0.935,0.945},
    xticklabels={(1,1), (2,2), (3,3), (4,4), (5,5), (6,6), (7,7)},
    ymajorgrids=true,
    grid style=dashed,
    tick label style={font=\tiny},
    label style={font=\tiny},
    title style={font=\footnotesize,align=center, yshift=-0.37\columnwidth},
]

\addplot[
    dashed,
    color=red,
    mark=diamond,
    mark options=solid,
    ]
    coordinates {
    (0,0.9024)(1,0.9181)(2,0.9218)(3,0.9367)(4,0.9458)(5,0.9441)(6,0.9445)
    };

\end{axis}
\end{tikzpicture}
\begin{tikzpicture}
\begin{axis}[
    title = {(b) Registration Accuracy},
    outer sep=0,
    width=0.57\columnwidth,
    height=0.42\columnwidth,
    xlabel={Number of stages: (Extraction, Registration)},
x label style={at={(axis description cs:0.5,0.17)},anchor=north},
y label style={at={(axis description cs:0.254,.5)},anchor=south},
    % xlabel near ticks,
    % ylabel near ticks,
    ylabel={Dice (Registration)},
    xmin=-0.3, xmax=6.30,
    ymin=0.5626, ymax=0.630,
    xtick={0,1,2,3,4,5,6},
    ytick={0.565,0.575, 0.585, 0.595, 0.605, 0.615, 0.625},
    yticklabels={0.565, 0.575, 0.585, 0.595, 0.605, 0.615, 0.625},
    xticklabels={(1,1), (2,2), (3,3), (4,4), (5,5), (6,6), (7,7)},
    ymajorgrids=true,
    grid style=dashed,
    tick label style={font=\tiny},
    label style={font=\tiny},
    title style={font=\footnotesize,align=center, yshift=-0.37\columnwidth},
]

\addplot[
    dashed,
    color=blue,
    mark=triangle,
    mark options=solid,
    ]
    coordinates {
    (0,0.5656)(1,0.5835)(2,0.5841)(3,0.6141)(4,0.6261)(5,0.6263)(6,0.6262)
    };

\end{axis}
\end{tikzpicture}
    \vspace{-11pt}
    \caption{Performance of ERNet with different number of extraction and registration stages.}
    \label{fig:stage dice}
    \vspace{-11pt}
\end{figure}

\subsubsection{Influence of Parameters.}
We study three important hyperparameters of our ERNet, \ie the number of stages for extraction and registration, the value of mask smoothing regularization parameter~$\lambda$ and the value of Sigmoid function slope parameter~$\gamma$.

\noindent\textbf{Number of stages of extraction and registration.} In our multi-stage design, the number of stages corresponds to network depth and the number of repeated extraction and registration.
In other words, more stages mean more refinements of the stripping and alignment, which is usually beneficial to the improvement of the extraction and registration accuracy.
As shown in Table~\ref{tab: stage res}, the ablation study demonstrates that both the extraction network and the registration network of ERNet are essential, and removing either one of them causes invalid results.
As illustrated in Figure~\ref{fig:stage dice}, we vary the number of stages of extraction and registration to learn their effects.
The results indicate that the performance of extraction and alignment improves with additional stages.
This supports the idea that a multi-stage design yields improved overall performance.

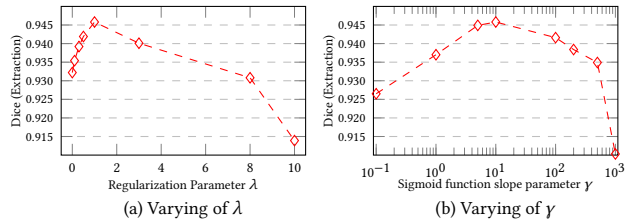
\begin{figure}[t]
\begin{tikzpicture}
\begin{axis}[
    title = {(a) Varying of $\lambda$},
    outer sep=0,
    width=0.57\columnwidth,
    height=0.42\columnwidth,
    xlabel={Regularization Parameter $\lambda$},
x label style={at={(axis description cs:0.5,0.16)},anchor=north},
y label style={at={(axis description cs:0.254,.5)},anchor=south},
    % xlabel near ticks,
    % ylabel near ticks,
    ylabel={Dice (Extraction)},
    xmin=-0.5, xmax=10.5,
    ymin=0.910, ymax=0.950,
    xtick={0,2,4,6,8,10},
    ytick={0.915, 0.920, 0.925, 0.930, 0.935,0.940, 0.945},
    yticklabels={0.915, 0.920, 0.925, 0.930, 0.935,0.940, 0.945},
    ymajorgrids=true,
    grid style=dashed,
    tick label style={font=\tiny},
    label style={font=\tiny},
    title style={font=\footnotesize,align=center, yshift=-0.37\columnwidth},
]

\addplot[
    dashed,
    color=red,
    mark=diamond,
    mark options=solid,
    ]
    coordinates {
    (0,0.9322)(0.1,0.9354)(0.3,0.9392)(0.5,0.9419)(1.0,0.9458)(3.0,0.9401)(8.0,0.9308)(10.0,0.9139)
    };

\end{axis}
\end{tikzpicture}
\begin{tikzpicture}
\begin{semilogxaxis}[
    %log ticks with fixed point,
    % extra x tick style={
    % log identify minor tick positions=true,
    % },
    % inner sep=0,
    title = {(b) Varying of $\gamma$},
    outer sep=0,
    width=0.57\columnwidth,
    height=0.42\columnwidth,
    xlabel={Sigmoid function slope parameter $\gamma$},
x label style={at={(axis description cs:0.5,0.16)},anchor=north},
y label style={at={(axis description cs:0.254,.5)},anchor=south},
    ylabel={Dice (Extraction)},
    xmin=0.09, xmax=1100,
    ymin=0.910, ymax=0.950,
    xtick={0.1, 1, 10, 100, 1000},
    ytick={0.915, 0.920, 0.925, 0.930, 0.935,0.940, 0.945},
    yticklabels={0.915, 0.920, 0.925, 0.930, 0.935,0.940, 0.945},
    ymajorgrids=true,
    grid style=dashed,
    tick label style={font=\tiny},
    label style={font=\tiny},
    title style={font=\footnotesize,align=center, yshift=-0.37\columnwidth},
]

\addplot[
    dashed,
    color=red,
    mark=diamond,
    mark options=solid,
    ]
    coordinates {
    (0.1,0.9265)(1,0.9370)(5,0.9449)(10,0.9458)(100,0.9416)(200,0.9384)(500,0.9349)(1000,0.9103)
    };

\end{semilogxaxis}
\end{tikzpicture}
    \vspace{-12pt}
     \caption{Effect of varying the mask smoothing regularization parameter $\lambda$ and sigmoid function slope parameter $\gamma$.}
    \label{fig:lambda}
    \vspace{-12pt}
\end{figure}

\noindent\textbf{Mask smoothing parameter $\lambda$.} As mentioned in Section~\ref{section:end-to-end training}, we introduce an regularization term to smooth the predicted masks.
To show the effectiveness of the smoothing regularization, we vary different values of the smoothing parameter $\lambda$ as shown in Figure~\ref{fig:lambda}(a).
The optimal Dice score occurs when $\lambda$ =1, while the performance gets worse as the $\lambda$ increases more or decreases.
This indicates that our ERNet benefits from mask smoothing regularization. 

\noindent\textbf{Sigmoid function slope parameter $\gamma$.} To show the effectiveness of the shifted sigmoid function we introduced in Eq.~(\ref{eq:sigmoid}), we evaluate the performance of our model under different $\gamma$ settings in a range from $10^{-1}$ to $10^3$. As shown in Figure~\ref{fig:lambda}(b), the model achieves the best performance when $\gamma$ = $10^{1}$, which is better than using the standard sigmoid function ($\gamma$ = $10^{0}$). As the $\gamma$ increases to $10^{2}$ and $10^{3}$, the dice score drops significantly because the large flat region of the sigmoid function prevents the error from backpropagation.

% \begin{figure}[t]
%   \centering
%   \includegraphics[width=0.75\linewidth]{fig/regularization_lambda_resized.pdf}
%   \vspace{-10pt}
%   \caption{Effect of varying the mask smoothing regularization parameter $\lambda$ on Dice score of extraction task.}
%   \label{fig:lambda}
%   \vspace{-10pt}
% \end{figure}

% \begin{figure}[t]
%   \centering
% \begin{tikzpicture}
% \begin{axis}[
%     width=0.84\columnwidth,
%     height=0.56\columnwidth,
%     xlabel={Regularization Parameter $\lambda$},
%     ylabel={Dice (Extraction)},
%     xmin=-0.5, xmax=10.5,
%     ymin=0.912, ymax=0.947,
%     xtick={0,2,4,6,8,10},
%     ytick={0.915,0.920,0.925,0.930,0.935,0.940,0.945},
%     yticklabels={0.915,0.920,0.925,0.930,0.935,0.940,0.945},
%     ymajorgrids=true,
%     grid style=dashed,
%     tick label style={font=\small},
%     label style={font=\normalsize},
% ]

% \addplot[
%     dashed,
%     color=red,
%     mark=o,
%     mark options=solid,
%     ]
%     coordinates {
%     (0,0.9322)(0.1,0.9354)(0.3,0.9392)(0.5,0.9419)(1.0,0.9458)(3.0,0.9401)(8.0,0.9308)(10.0,0.9139)
%     };

% \end{axis}
% \end{tikzpicture}
%   \vspace{-10pt}
%   \caption{Effect of varying the mask smoothing regularization parameter $\lambda$ on Dice score of extraction task.}
% %   \label{fig:lambda}
%   \vspace{-10pt}
% %\label{fig:lambda}
% \end{figure}
\section{Related Work}
\label{sec:related}
\noindent \textbf{Brain extraction.} Over the past decade, myriad methods have been proposed, emphasizing the importance of the brain extraction problem.
% In general, brain regions manually outlined by human experts in raw MRI images are considered as ground truth; however, this is time-consuming, label-intensive, and impractical to perform on a large scale.
Smith et al.~\cite{smith2002fast} proposed a deformable model to fit the brain surface using a locally adaptive set model.
3dSkullStrip~\cite{cox1996afni} is a modified version of~\cite{smith2002fast}, which uses points lying outside the brain surface to guide the evolution of the mesh.
Shattuck et al.~\cite{shattuck2002brainsuite} employs anisotropic diffusion filtering and a 2D Marr Hildreth edge detector to identify the brain boundary.
% Apart from methods, several other traditional approaches~\cite{segonne2004hybrid,iglesias2011robust,eskildsen2012beast,lutkenhoff2014optimized} are also commonly used for brain extraction.
Apart from methods, several other traditional approaches~\cite{segonne2004hybrid,iglesias2011robust,eskildsen2012beast} are also commonly used for brain extraction.
However, the aforementioned methods rely heavily on parameter setting and manual quality control, which are time-consuming and labor-intensive. 
Recently, deep learning-based methods have been introduced for brain extraction due to their superior capability and extreme speed.
Kleesiek et al.~\cite{kleesiek2016deep} proposed a voxel-wise 3D CNN for skull stripping.
%Lucena et al.~\cite{lucena2019convolutional} introduced data augmentation and used silver standard masks to make the network robust.
Hwang et al.~\cite{hwang20193d} suggested that 3D-UNet yields highly competitive results for skull stripping. 
%Most recently, Zhong et al.~\cite{zhong2021dika} used Dual Self-Attention Module (DSAM) to fuse the age-specific intensity information and domain-invariant prior knowledge for guided skull stripping. 
The above learning-based methods often demand a large amount of adequately labeled data for effective training.
However, neuroimage datasets are usually small and expensive to annotate.

\noindent \textbf{Image registration.}
%Traditional image registration methods~\cite{avants2009advanced, bajcsy1989multiresolution, avants2008symmetric, jenkinson2001global, saad2009new, schnabel2001generic} often try to maximize the similarity between images by iteratively optimizing the transformation parameters, where mean square error (MSE), normalized cross-correlation (NCC) and mutual information (MI), \etc, are commonly used as intensity-based similarity measures.
Traditional image registration methods~\cite{avants2009advanced, avants2008symmetric, jenkinson2001global, saad2009new} often try to maximize the similarity between images by iteratively optimizing the transformation parameters, where normalized cross-correlation
(NCC) and mutual information (MI), \etc, are commonly used as intensity-based similarity measures.
However, iteratively optimizing each pair of images tends to face the drawbacks of high computational cost and being trapped in local optima, resulting in failing to yield an efficient and robust registration result.
Recently, many deep learning-based methods have been proposed due to their superior computational efficiency and registration performance.
Sokooti et al.~\cite{sokooti2017nonrigid} proposed a multi-scale 3D CNN termed as RegNet to learn the artificially generated displacement vector field (DVF) for 3D chest CT registration.
%Dai et al.~\cite{dai2020dual} introduced dual-attention recurrent networks for neuroimaging data registration. 
While these methods present competitive results, they are all supervised.
In other words, the training procedure is guided by ground truth transformations. In practice, obtaining high-quality ground truth is often expensive in medical imaging.
To address this limitation, unsupervised registration methods~\cite{balakrishnan2018unsupervised,zhao2019recursive} received much attention and delivered promising results.
However, the above methods rely on accurate skull stripping results to perform the registration, where manual visual inspection is required to remove the inaccurate extracted image.
Such human involvement is not only time-consuming but also brings in biases.
Unlike these works, we pursue unsupervised joint learning for extraction and registration.

% %-----------------------------------------------
% % Conclusion Section
\section{Conclusion}
\label{sec:con}

In this paper, we propose a novel unified end-to-end framework, called ERNet, for unsupervised collective extraction and registration.
Different from previous work, our proposed method seamlessly integrated two tasks into one system to achieve joint optimization.
Specifically, ERNet contains a pair of multi-stage extraction and registration modules.
These two modules help each other boost extraction and registration performance simultaneously without any annotation information.
Moreover, the multi-stage design allows each task to proceed incrementally, thus refining their respective performance to a better extent.
The experimental results demonstrate that ERNet not only outperforms state-of-the-art approaches in both extraction and registration accuracy but is also more robust and time-efficient.

\section{Acknowledgments}
\label{sec:ack}
Lifang He was supported by Lehigh's accelerator grant S00010293.

%\newpage
% balance the two columns of the last page
%\balance

\bibliographystyle{ACM-Reference-Format}
\bibliography{08_reference}

%\newpage
\appendix
\section{Appendix for reproducibility}
\label{sec:appendix}

%This section provides more details on dataset preprocessing, ERNet settings and baseline settings to support the reproducibility of the results in this paper. We have released our code and data publicly available at~\url{https://github.com/ERNetERNet/ERNet}.

This section provides more details to support the reproducibility of the results in this paper. We have released our code and data publicly available at~\url{https://github.com/ERNetERNet/ERNet}.

\subsection{Details of Data Preprocessing}
We evaluate the effectiveness of our proposed method on three different public brain MRI datasets, LPBA40, CC-359 and IBSR. 
%Each dataset is split into training, validation and test sets, respectively. We apply intensity normalization to each scan and rescale voxel values to a range between zero and one.
Table~\ref{tab:dataset} summarizes the properties of the datasets.

\begin{table}[h]
    \centering
    \vspace{-8pt}
    \caption{Summary of datasets}
    \label{tab:dataset}
    \vspace{-9pt}
    \resizebox{0.88\linewidth}{!}{
    \begin{tabular}{lccc}
    \toprule
     & \textbf{LPBA40}& \textbf{CC359 }& \textbf{IBSR}\\
    \midrule
    Raw size  & $256\times124\times256$ & $171\times256\times256$ & $256\times256\times128$\\
    Cropped size  & $96\times96\times96$ & $96\times96\times96$ & $96\times96\times96$\\
    Training & 30  & 298 & -\\
    Validation & 5  & 30 & -\\
    Test & 4  & 30 & 18\\
    \bottomrule
    \end{tabular}}
    \vspace{-8pt}
\end{table}

\noindent\textbullet\ \textit{LONI Probabilistic Brain Atlas (LPBA40)}~\cite{shattuck2008construction}: 
This dataset consists of 40 raw T1-weighted 3D brain MRI scans (40 different patients) along with their brain masks and the corresponding segmentation ground truth of 56 anatomical structures. The brain mask and anatomical segmentations are used to evaluate the accuracy of extraction and registration, respectively.
Same to~\cite{balakrishnan2018unsupervised, zhao2019recursive}, we focus on atlas-based registration, where the first scan is the target image and the remaining scans need to align with it. Among the 39 scans, we use 30, 5, and 4 scans for training, validation, and test, respectively. All scans are resized to $96\times96\times96$ after cropping.

\noindent\textbullet\ \textit{Calgary-Campinas-359 (CC-359)}~\cite{souza2018open}: 
This dataset consists of 359 raw T1-weighted 3D brain MRI scans (359 different patients) and the corresponding brain masks. It also contains the labeled white matter as the ground truth. We use the brain masks and white-matter masks to evaluate the accuracy of extraction and registration, respectively. Same to LPBA40, we concentrate on atlas-based registration and split CC359 into 298, 30, and 30 scans for training, validation, and test sets. All scans are resized to $96\times96\times96$ after cropping.

\noindent\textbullet\ \textit{Internet Brain Segmentation Repository (IBSR)} \cite{rohlfing2011image}: 
This dataset provides 18 raw T1-weighted 3D brain MRI scans (18 different patients) along with the corresponding segmentation results. 
We merge all segmentation results to construct the brain mask. 
Due to the small sample size, We use this dataset only to test the model trained on CC359. 
%Similarly, we follow atlas-based registration that all 18 scans need to align with the first scan of CC359. All scans are resized to $96\times96\times96$ after cropping.
Thus, all 18 scans need to align with the first scan of CC359. All scans are resized to $96\times96\times96$ after cropping.

\subsection{Details Settings of ERNet}
\textbf{Training settings of ERNet.} Our experiments are running on Red Hat Enterprise Linux 7.3 with an Intel$^{\circledR}$ Xeon$^{\circledR}$ E5-2667 v4 CPU and an NVIDIA Tesla V100 GPU. 
%The code released is in Python 3.7.6 and the implementation of neural networks is based on PyTorch 1.7.1. We also employ Numpy 1.21.6, SimpleITK 2.0.2 and Nibabel 3.1.1 in the implementation. 
The implementation of neural networks is based on PyTorch 1.7.1.
During the training process, we apply batch gradient descent with each training batch consists of one pair of images to address GPU memory limitation. Models are optimized using Adam optimizer~\cite{kingma2014adam} with a learning rate of $1 \times 10^{-6}$. We also leverage the image augmentation technique to expand the datasets, where a transformation with random translation, rotation, and scaling is applied to source images. We detail it in Table~\ref{tab:transformation}.

%%%%% No time %%%%%%%%%%%
\begin{table}[h]
    \centering
    \vspace{-2pt}
    \caption{Range of random transformation.}
    \label{tab:transformation}
    \vspace{-8pt}
    \resizebox{0.68\linewidth}{!}{
    \begin{tabular}{lccc}
    \toprule
    
    \multicolumn{1}{c}{\multirow{3}{*}{Datasets}} &   \multicolumn{3}{c}{Transformation} \\

    \cmidrule(lr){2-4}
    
        &   Translation  & Rotation       & Scale         \\
        &   (Voxels)  & (Degree)  & (Times)    \\

    \midrule
    LPBA40 & $\pm$ 5 & $\pm$ 5  & 0.98 $\sim$ 1.02\\
    
    CC359 & $\pm$ 3 & $\pm$ 3  & 0.99 $\sim$ 1.01\\

    \bottomrule
    \end{tabular}}
    \vspace{-13pt}
\end{table}

\noindent\textbf{Parameters settings of ERNet.} Since we design a multi-stage extraction and registration network, the extraction and registration stages are both set to 5 in this work. The best mask smoothing parameter $\lambda$ in Eq.~(\ref{eq:loss}) is 1, and the best sigmoid function slope parameter $\gamma$ in Eq.~(\ref{eq:sigmoid}) is $10^{1}$.
%The extraction network takes a 3D U-Net as the backbone with an input size of $2\times96\times96\times96$ and an output size of $1\times96\times96\times96$.
The extraction network contains 10 convolutional layers with 16, 32, 32, 64, 64, 64, 32, 32, 32 and 16 filters. 
The registration network adopt 3D CNNs and fully-connected layers to map the input to the dimension of $1\times12$. It contains 6 convolutional layers with 16, 32, 64, 128, 256 and 512 filters. The output dimensions of the 2 fully-connected layers are 128 and 12.

\newcommand{\code}[1]{\texttt{#1}}

\subsection{Settings of Baselines}

\noindent\textbf{Brain Extraction Tool (BET)} \cite{smith2002fast}: This is a skull stripping method included in FSL package. It uses a deformable approach to fit the brain surface by applying locally adaptive set models. The command we use for BET is \code{bet <input> <output> -f 0.5 -g 0 -m}, where \texttt{f} and \texttt{g} are fractional intensity threshold and gradient in fractional intensity threshold, respectively. We set them to default values.

\noindent\textbf{3dSkullStrip}~\cite{cox1996afni}: This is a modified version of BET that is included in the AFNI package. It performs skull stripping based on the expansion paradigm of the spherical surface. The command we use for 3dSkullStrip is \code{3dSkullStrip -input <input>  -prefix <output> -mask\_vol -fac 1000}. \code{fac} is set to the default value.

\noindent\textbf{Brain Surface Extractor (BSE)}~\cite{shattuck2002brainsuite}: It extracts the brain region based on morphological operations and edge detection, which employs anisotropic diffusion filtering and a Marr Hildreth edge detector for brain boundary identification. The command we use for BSE is \code{bse -i <input> -o <output> --mask <mask> -p --trim --auto --timer }. Hyperparameters are set to default values.

\noindent\textbf{FMRIB's Linear Image Registration Tool (FLIRT)}~\cite{jenkinson2001global}: This is a fully automated affine brain image registration tool in FSL package. The command we use for FLIRT is \code{flirt -in <source> -ref <target> -out <output> -omat <output parameter> -bins 256 -cost corratio -searchrx -90 90 -searchry -90 90 -searchrz -90 90 -dof 12  -interp trilinear}.

\noindent\textbf{Advanced Normalization Tools (ANTs)} \cite{avants2009advanced}: It is a state-of-the-art medical image registration toolkit. Here we utilize affine transformation model and cross-correlation metric for registration. 

\noindent\textbf{VoxelMorph (VM)} \cite{balakrishnan2018unsupervised}: This unsupervised, deformable image registration method employs a neural network to predict the nonlinear transformation between images. 
For network architectures, we use the latest version, VoxelMorph-2, and configure 10 convolutional layers with 16, 32, 32, 64, 64, 64, 32, 32, 32 and 16 filters. 
%The kernel size of each convolutional layer is 3 × 3.
The ratio of deformation regularization is set to 10.

\noindent\textbf{Cascaded Registration Networks (CRN)} \cite{zhao2019recursive}:
It is an unsupervised multi-stage registration method. In different stages, the source image is repeatedly deformed to align with a target image. Same to ERNet, the number of stages is set to 5. In each stage, we configure 10 convolutional layers with 16, 32, 32, 64, 64, 64, 32, 32, 32 and 16 filters. 
%The kernel size of each convolutional layer is 3 × 3. 
The ratio of deformation regularization is set to 10.

\noindent\textbf{ERNet w/o Ext}: This is a variant of ERNet where we remove the extraction modules. Here it is a registration method only.

\end{document}